\documentclass[lettersize,journal]{IEEEtran}

\usepackage{algorithmic}
\usepackage{algorithm}
\usepackage{array}
\usepackage[caption=false,font=normalsize,labelfont=sf,textfont=sf]{subfig}
\usepackage{textcomp}
\usepackage{stfloats}
\usepackage{verbatim}
\usepackage{graphicx}
\usepackage{cite}
\usepackage{color}

\PassOptionsToPackage{hyphens}{url}
\usepackage{url}
\usepackage{amsmath,amssymb,amsfonts}
\usepackage{booktabs}
\usepackage{textcomp}
\usepackage{xcolor}
\usepackage{soul}
\usepackage{caption}
\usepackage[export]{adjustbox}
\usepackage[normalem]{ulem}

\graphicspath{{Images/}} 
\newcommand{\quotes}[1]{``#1''}

\begin{document}

\title{BUDD-e: an autonomous robotic guide for visually impaired users}

\author{Jinyang Li, Marcello Farina, Luca Mozzarelli, Luca Cattaneo, Panita Rattamasanaprapai, Eleonora A. Tagarelli, Matteo Corno,\\
	Paolo Perego, Giuseppe Andreoni, Emanuele Lettieri
\thanks{Jinyang Li, Marcello Farina, Luca Mozzarelli, Luca Cattaneo, Panita Rattamasanaprapai, Eleonora A. Tagarelli, and Matteo Corno are with the Dipartimento di Elettronica, Informazione e Bioingegneria, Politecnico di Milano, Via Ponzio 34/5, Milan, Italy (email: jinyang.li@polimi.it, marcello.farina@polimi.it, luca.mozzarelli@polimi.it, luca.cattaneo@polimi.it, panita.rattamasanaprapai@mail.polimi.it, eleonoraantonia.tagarelli@mail.polimi.it,  matteo.corno@polimi.it); Paolo Perego and Giuseppe Andreoni are with the Department of Design, Politecnico di Milano, Via Durando, Milan, Italy (email: paolo.perego@polimi.it and giuseppe.andreoni@polimi.it); Emanuele Lettieri is with the Department of Management, Economics and Industrial Engineering, Politecnico di Milano, Via Lambruschini 4, Milan, Italy (e-mail: emanuele.lettieri@polimi.it).}
\thanks{This work has been financially supported by the Politecnico di Milano through the Polisocial Award 2021 project Blind-assistive aUtonomous Droid Device (BUDD-e), by the PRIN 2022 project Control of Assistive Robots in crowded Environments (CARE, Id. 20225LX9M3), and by the Italian Ministry of Enterprises through the project 4D Drone Swarms (4DDS) under grant no. F/310097/01-04/X56.}}



\maketitle

\begin{abstract}
This paper describes the design and the realization of a prototype of the novel guide robot BUDD-e for visually impaired users. The robot has been tested in a real scenario with the help of visually disabled volunteers at ASST Grande Ospedale Metropolitano Niguarda, in Milan. The results of the experimental campaign are throughly described in the paper, displaying its remarkable performance and user-acceptance.
\end{abstract}


\section{Introduction}
According to \cite{Abstract20}, in 2020 the number of
totally blind people was estimated to about 49.1 million (about 0.6 \% of the world population), while people with severe and moderate vision problems were estimated to 33.6 million (about 0.4 \% of the world population) and 221.4 million (about 2.8 \% of the world population), respectively. Furthermore, due to an aging population,
it is estimated that the rate of people affected by vision
problems will continue to increase in the coming decades \cite{Ageing21}.\\
\noindent
People with visual impairments currently face a number of issues when it comes to visiting public spaces and using services. It is very difficult for blind and partially sighted persons to access shared places (areas where cars, buses, pedestrians, and cyclists share the same space) alone since important inclusive environmental aids are frequently removed in communal areas.
As discussed in~\cite{inclusivecitymaker}, navigating inside a shopping mall for a blind or low-vision person can be tiring and stressful.
Shopping in groceries is practically impossible and shopping centers often don't have enough staff on duty to offer help.
Even finding and getting into shops can be problematic, also due to the fact that assistance dogs often experience access refusals.
According to~\cite{accident}, visual loss increases the likelihood of unintentional injury and there is a strong link between the type, severity, and number of injuries and the degree of vision loss.
Consequently, as stated in~\cite{OutdoorMobility}, at present a large number of blind and low-vision people require sighted companions while traveling outside.
\noindent
Similar issues are experienced in hospitals and medical centers and by sports practitioners.
For instance, for visually impaired persons, in order to practice sport (e.g., running), continuous assistance and the presence of guides are presently strictly necessary.

\noindent
A number of assistive technologies have been introduced in the past years to make it easier for visually impaired people to benefit from services, visit shopping centers, groceries, and hospitals, and practice sports.
As discussed in~\cite{Guerreiro}, GPS-based navigation systems are probably the most widespread systems used for outdoor environments.
In indoor scenarios, the most widespread technologies span from the ones making use of smartphone sensors to camera-based systems (where a remote guide gives instructions to the users), to WiFi or Bluetooth Low-Energy beacons.

\noindent
Besides the above-mentioned localization and tracking technologies, there has been also a growing research dedicated to the development of enhanced canes and guide robots.

\noindent
Regarding the former class, a number of enhanced and “virtual” canes have been developed, with different features, e.g., Sonic Torch, Pathsounder, Ultracane, Teletact~\cite{EyeCane}.
Recent systems embed sensors (e.g., cameras and/or infra-red sensors) capable of providing information on the ground and/or on the presence of nearby obstacles and provide tactile and/or acoustic feedbacks to the user, e.g., the Mygo Cane~\cite{Veeramachaneni}, the Eyecane~\cite{EyeCane}, BBEEP~\cite{BBeep}, the Guide Cane~\cite{614314}, the Glide~\cite{glide} and the Co-robotic cane~\cite{7549220}.

\noindent
On the other hand, robot guides are devised to mimic the functions of traditional navigation aids while overcoming some of their limitations. They are used either in front or on the side of the user (mimicking guide dogs or sighted guides).

\noindent
To the best of the authors' knowledge, the first proposal in this class was a robot-assisted wayfinding system described in~\cite{DBLP:journals/arobots/KulyukinGNO06}, consisting of a mobile robotic guide whose architecture presents a path planner, a behavior manager, and a user interface (UI).
The navigational part is supported by small passive RFID sensors embedded in the known indoor environment to be visited, while the UI has both haptic and speech inputs.

\noindent
Other solutions have been proposed by providing walkers with navigation functionalities. For instance, the c-Walker robot~\cite{dali}, in case the user takes a wrong path or in case of obstacles to avoid, intervenes passively generating haptic and acoustic signals or actively taking control of the movement. Similarly, the Smart Walker~\cite{smartwalker} communicates with the user by means of two vibrating handles and a belt worn by the user around the waist.

\noindent
Solutions based on available mobile robotic platforms have also been proposed.
For example, in~\cite{Kulkarni} the prototype of a guide robot based on the Pioneer 3DX platform and endowed with a D handle is proposed, including a few interactive features, e.g., sound generation for identifying the robot's location and a speech emitter for communication with the user.
In~\cite{Tobita} a robot (to be deployed in hospitals, government centers, etc.) is proposed embedding collision avoidance capabilities, but which receives moving instructions by the user.
In~\cite{HepticRein} a novel haptic rein is designed to support non-visual and non-auditive communication between a visually impaired user and a robot guide, based on force sensors and vibrating actuators. A similar solution is the Robotic Guide Dog described in~\cite{xiao2021robotic} based on the Mini Cheetah~\cite{8793865}, and connected to the human by a leash.
With the use of a force sensor, the guide dog pulls and loosens the rope to bring the user to the final location goal.
In~\cite{Chuang} the authors realized a robotic guide capable of autonomously recognizing and following different man-made trails.
In~\cite{Cabot} the CaBot (Carry-on roBot) is described, i.e., an autonomous suitcase-shaped navigation robot that is able to guide blind users to a destination while avoiding obstacles on their path.
Also, the Ballbot~\cite{8968546} has been proposed as a tool for physically leading people: it is an indoor service robot that provides people with physical assistance and active guidance.
The framework combines a planning algorithm and a human-robot interaction module to guide the person to a specified planned position.

\noindent
In our opinion, however, so far the existing solutions are not capable of embedding, at the same time, the following functionalities: (i) the capability of navigating in both indoor and outdoor possibly populated environments, while following defined tracks and avoiding collisions; (ii) the capability of smoothly adapting to the user's velocity, including fast walking (e.g., in parks) and running (e.g., in sports centers); (iii) the capability of providing, in an adaptive and constant fashion, physical commands to the user, robustly with respect to the user motion.
The latter, in particular, has been highlighted as a fundamental feature in surveys conducted in the last years~\cite{bigsurvey,SurveyBUDD-e}. In particular, existing solutions lack a suitable system allowing for active interaction between the user and the robot, making it capable, e.g., through the rod and a handle, to exert a constant force to the user irrespective of the velocity variations of the user and of the robot.

\noindent
This work has been conducted in the framework of the project BUDD-e\footnote{For information see \url{https://budd-e.polimi.it/}} (Blind-assistive aUtonomous Droid Device), whose goal is to improve the quality of life of blind and partially-sighted persons by providing accessibility to services and public spaces~\cite{Rebecchi_Farina_Andreoni_Capolongo_Corno_Perego_Lettieri_2023}.
This goal is pursued by devising an ad-hoc robotic guide, to be deployed in structured spaces, e.g., sports, healthcare, shopping, and cultural centers.\\
In this paper we describe the design, the realization, and the testing phases of the novel guide robot BUDD-e. In particular, the robot has been tested in a real scenario with the help of visually impaired volunteers displaying promising performance and results.\smallskip\\
The paper is structured as follows. In Section~\ref{sec:structure} we describe the structure of BUDD-e, as well as the control algorithms which are designed and used for adapting the robot speed to the user's motion. In Section~\ref{sec:autonomous_driving} we provide details on the main components of BUDD-e's autonomous driving stack. Then, in Section~\ref{sec:validation} we describe the experimantal campaign carried out at ASST Grande Ospedale Niguarda in Milan and the achieved results. Finally, in Section~\ref{sec:conclusions} we draw some conclusions and we pave the way for the future research on the topic.
\section{The BUDD-e robotic guide}
\label{sec:structure}
In this section, we first describe the mechanical design of BUDD-e. The section also presents the general control architecture and the target tracking scheme used by BUDD-e.

\subsection{The mechanical setup}
\label{se:mech-design}
\begin{figure}[H]
	\centering
	\includegraphics[width = 0.8\columnwidth]{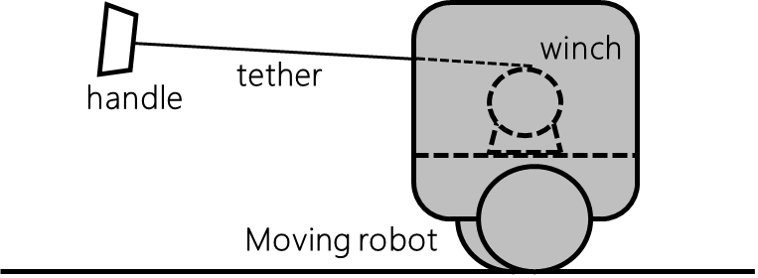}
	\caption{Scheme of the BUDD-e robotic guide}
	\label{fig:entire_system_model}
\end{figure}

The whole setup is roughly illustrated in Figure~\ref{fig:entire_system_model}, consisting of a robotic platform, a winch, and a tether, which has a non-extensible part and an elastic one, connected to the handle.

\begin{figure}[h!]
	\centering
	\includegraphics[width=0.4\columnwidth]{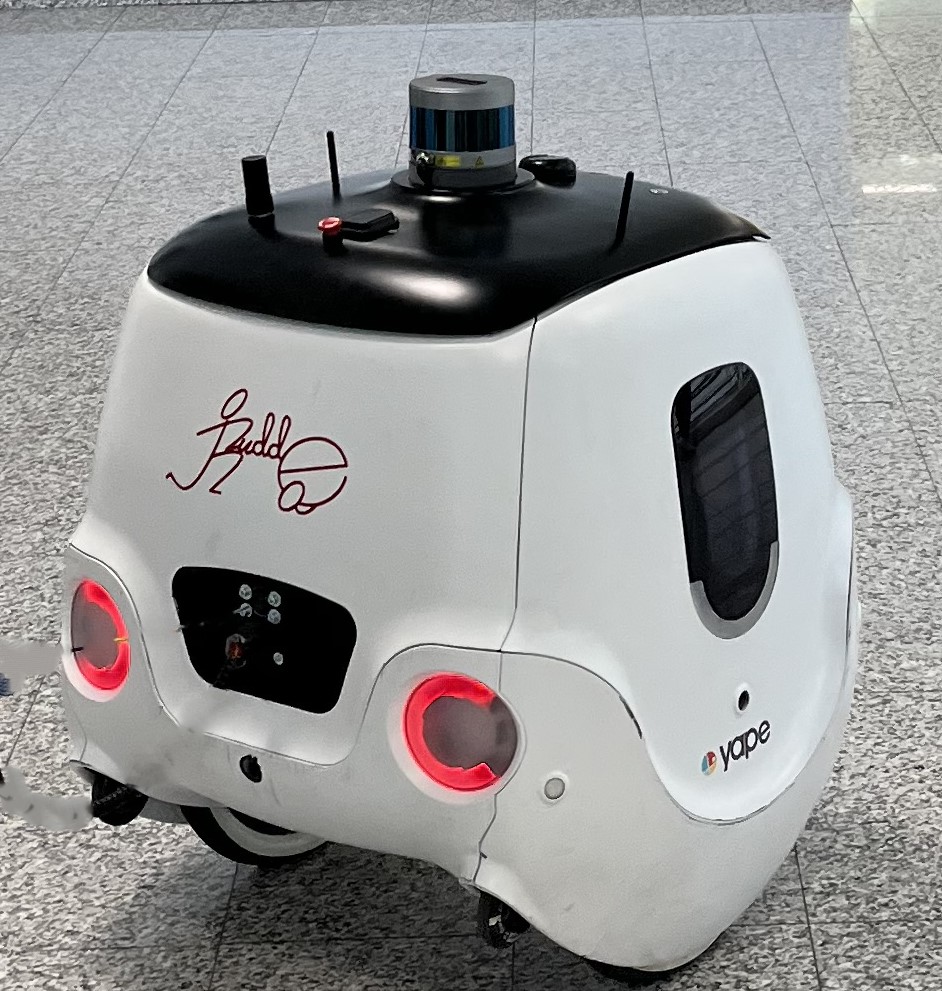}
	\caption{The Yape robot}
	\label{fig:yape}
\end{figure}

\noindent
In the BUDD-e project, the robotic platform is Yape~\cite{Yape2016}, a self-balancing autonomous two-wheeled robot currently used for last-mile delivery uses, see Figure~\ref{fig:yape}, produced by Yape S.r.l. 
Yape is a differential drive mobile robot with a Two-Wheeled Inverted Pendulum (TWIP) structure.
The torque needed for stabilization and driving purposes is given by two brushless DC motors housed in the hub of the two bicycle-like wheels.
All the components are contained in a steel chassis with a 60 cm wide cubic structure, covered with high-density polystyrene, with a total mass of around 40 kg.
The robot can reach a maximum velocity of about 20 km/h and can handle slopes up to 10°. 
In Yape the stabilization, speed control, trajectory planning, and tracking are guaranteed by the use of both proprioceptive and exteroceptive sensors.
Specifically, an Inertial Measurement Unit (IMU) with 6 degrees of freedom (3-axis accelerometer and a 3-axis gyroscope) is attached to the inner structure, while the RS16 Robosense 16-layer LiDAR sensor is mounted on the top of the vehicle.
Data acquired from the proprioceptive sensors are managed by the Vehicle Control Unit (VCU), a central microcontroller tasked with running the stabilization and speed tracking algorithms~\cite{parravicini_robust_2019}.
Torque references are sent to the wheel motors by means of a CAN bus line.
The robot is also equipped with a Navigation Control Unit (NCU): an Nvidia Xavier AGX board, which collects data from the exteroceptive sensors, runs the decision-making algorithms and communicates velocity references to the VCU.

\begin{figure}[h!]
	\centering
	\includegraphics[width=0.5\linewidth]{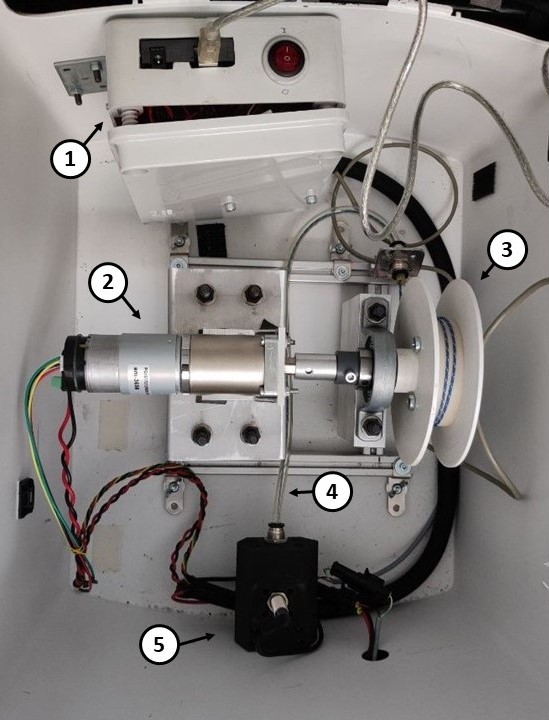}
	\includegraphics[width=0.9\linewidth]{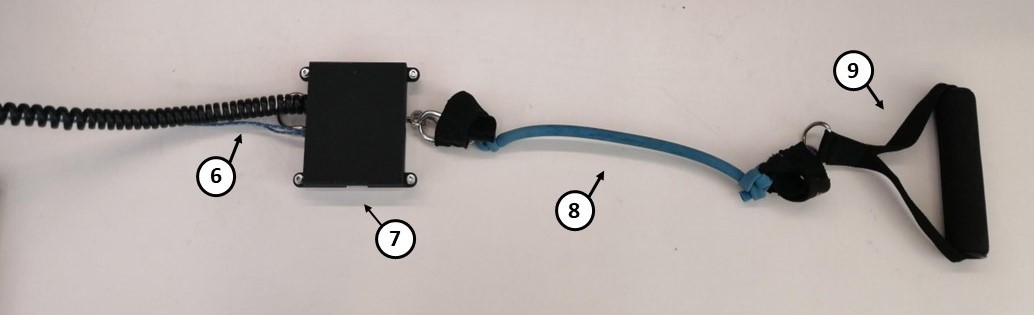}
	\caption{The Smart Tether System: 1. Box with electronic components; 2. Motor; 3. Shaft; 4. Polyurethane tube; 5. Proximity sensor; 6. Rigid cable; 7. Load cell; 8. Elastic cable; 9. Handle.}
	\label{fig:ST}
\end{figure}

\noindent
The ensemble composed of the winch and the tether (see Figure~\ref{fig:entire_system_model}) is denoted Smart Tether System.
The Smart Tether System, whose first design is described in detail in~\cite{STpaper}, is a custom-made add-on device designed specifically for {BUDD-e}.
It is physically situated in the box, contained in Yape's payload compartment, originally designed for transporting packages, see Figure \ref{fig:ST} (top panel).
The latter is composed of a DC motor, powered directly from Yape, equipped with an incremental encoder and a gearbox.
The rotation of the motor makes the shaft connected to it move at a given angular velocity.
During the operation, a cable is wound or unwound on the reel according to the shaft rotation.
The cable passes through an 8x5.5 98 shore polyurethane tube and exits through a custom-made box where a proximity sensor is fixed.
The main role of the proximity sensor is to detect a specific part of the cable where a metallic wire is wrapped, updating the information of the exact position to the encoder.
In this way, the Smart Tether System can obtain a precise absolute cable position measure.
The logics are implemented in an Arduino board fixed inside a dedicated white box.
%
Outside of Yape, as depicted in Figure \ref{fig:ST} (bottom panel), the rigid cable is connected in series to a full-bridge 5 kg tension load cell and to an elastic tether, which is in turn connected to a handle.

\subsection{The control system}
\label{subsec:control_system}
As discussed, the objective of the overall control system is to safely guide a visually impaired user to a specific destination and, at the same time, to avoid moving obstacles along the path.
BUDD-e can be roughly represented as a control system with three degrees of freedom and three main controllers.

\noindent
For a smooth behavior and to ensure the safety of the user, a dedicated distance controller regulates the longitudinal dynamics of BUDD-e, maintaining the distance $d$ from the user as close as possible to a reference one $d^o$ and adapting its speed $v$ the user's velocity $v_{\rm VI}$.
The control input used is the reference longitudinal velocity of Yape $v_{\rm ref}$.
This module needs to be fed with a measure, or estimate, of the robot-user distance, which will be provided by a target tracking algorithm presented in Section \ref{subsec:target_tracking}.

\noindent
To ensure the interaction between the system and the user, the Smart Tether System is controlled by a dedicated force controller to provide the user with information about the correct path to follow.
This is done by maintaining the elastic force between the VI person and BUDD-e constant to allow the user to perceive the tension on the handle.
The control input used is the reference angular speed of the DC motor.
For a detailed presentation of the Smart Tether System model and control system design, the reader is deferred to~\cite{BUDDEpanita}.
Although in this work the set-up has been slightly changed with respect to the one in~\cite{BUDDEpanita}, the resulting model is analogous to the one identified and used in the previous work and the control design phase is the same.

\noindent
Eventually, a full autonomous driving stack needs to be implemented on the vehicle, including localization, perception, and global and local planning algorithms.
The latter module is of particular interest, aiming to control the longitudinal and lateral dynamics of BUDD-e to follow a pre-defined trajectory and avoid unexpected obstacles.
The control inputs used in this case are the BUDD-e's yaw rate and longitudinal velocity references: $\omega_{\rm ref}$ and $v_{\rm ref}$, respectively.
It is worth noting that both the distance controller and the local planner output a velocity reference: a suitable mixing logic should be developed to select the appropriate input based on the time-varying priorities.
The architecture of the autonomous driving stack will be detailed in Section \ref{sec:autonomous_driving}.

\subsection{The Yape longitudinal speed model identification}
\label{sec:switching_law}
The longitudinal velocity of Yape is controlled through a dedicated embedded control system, having $v_{\rm ref}$ as input and $v$ as output.
Therefore, in this work, we use $v_{\rm ref}$ as a manipulable input.
The distance control system, as described in Section~\ref{subsec:distance_controller}, has the role of maintaining the robot-user distance $d$ as close as possible to a reference value $d^0$ despite possibly random variations of the disturbance acting on the model, corresponding to the user's speed $v_{\rm VI}$, which can be estimated by the tracking target algorithm using LiDAR data, as described in Section~\ref{subsec:target_tracking}.

\begin{figure}[H]
	\centering
	\includegraphics[width=1\linewidth]{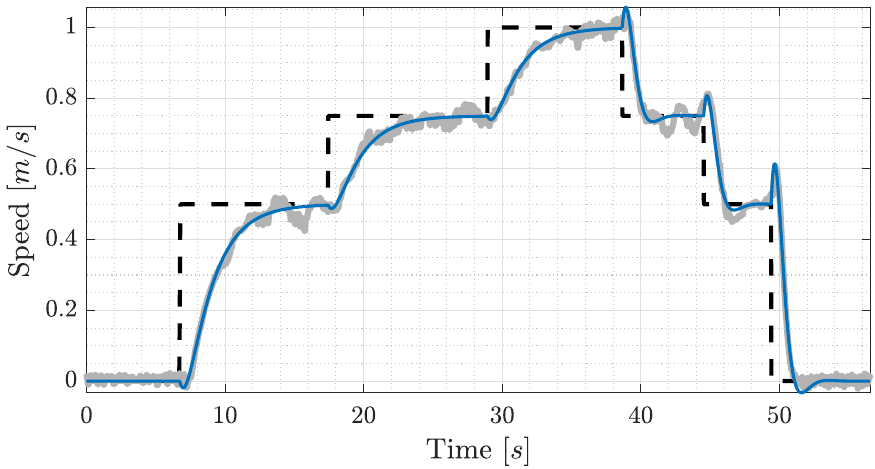}
	\caption{\label{fig:Yape_Identification}Example of the data used to identify the longitudinal dynamical model of Yape. Black dashed line: $v_{\rm ref}$; gray solid line: measured Yape velocity $v$; blue solid line: simulated Yape velocity $v$.}
\end{figure}

\noindent
The identification of the Yape dynamics was conducted starting from step experiments conducted at different speeds, see e.g., the data shown in Figure \ref{fig:Yape_Identification}.
These data reveal that Yape displays two different modes of operation, roughly depending on whether the robot accelerating or decelerating.
More specifically, when Yape is accelerating, its longitudinal dynamics can be represented by the transfer function $F_{\rm acc}(s)$, i.e.,
	\begin{equation}
		\label{acceleration_dynamic}
		F_{\rm acc}(s)=\frac{1-0.3423s}{1+2.3728s+0.9681s^2}
	\end{equation}
On the other hand, when Yape is decelerating, its longitudinal dynamics can be represented by the transfer function $F_{\rm dec}(s)$, i.e.,
	\begin{equation}
		\label{deceleration_dynamic}
		F_{\rm dec}(s)=\frac{1-0.4255s}{1+0.6187s+0.2059s^2}
	\end{equation}
Notice that both acceleration and deceleration modes are characterized by two asymptotically stable poles and a non minimum-phase zero.

\noindent
In Figure \ref{fig:Yape_Identification} we show the comparison of the real velocity with the one obtained by feeding the switching model with the input $v_{\rm ref}$ used in the experiment.
The switching condition used here is extremely simple (in view of the simplified input used in the experiment) and depends upon the step-wise input provided to the system.
Figure \ref{fig:Yape_Identification} shows that, when the commutation condition is correctly identified, the switching model is capable of capturing the system behavior precisely.
However, in the case of more complex input signals, it was not possible to precisely identify the switching condition between these two modes of operation.
In view of this, the implementation of a switching or time-varying control law could be critical; for this reason, in the next section we derive a time-invariant control law with LMI arguments, robust with respect to the specific mode of operation and to the switching between them.

\subsection{The Yape longitudinal speed discrete-time switching model}
For LMI-based control design, a discrete-time state-space realization of the system model will be derived.
To do so, we define $u(t)=a_{\rm ref}(t)=\dot{v}_{\rm ref}(t)$, used as new "virtual" input.
Also, we define the system state vector as ${x}(t) = [\:v_{\rm ref}(t) \:p(t)\: v(t)\: a(t)\: ]^T$, where $a(t)=\dot{v}(t)$ and $p(t)$ is the robot curvilinear abscissa.
Recalling that the transfer function between $v_{\rm ref}(t)$ and $v(t)$, for both $i={\rm acc},{\rm dec}$ has the form
\begin{equation}
	F_i(s) = \frac{1+sT_i}{1+\alpha_is+\beta_is^2}
\end{equation}
we can write
\begin{equation}
	\begin{cases}
		\label{eq:system_LMI}
		\dot{v}_{\rm ref}=u(t)\\
		\dot{p}(t)=v(t)\\
		\dot{v}(t)=a(t)\\
		\dot{a}(t) = \frac{1}{\beta_i}(-\alpha_i a(t)-v(t)+{v}_{\rm ref}(t)+T_i{a}_{\rm ref}(t))
	\end{cases}
\end{equation}
System \eqref{eq:system_LMI} can be written in compact form and discretized, with a sampling time $T_s=0.1$ s, obtaining
\begin{equation}
	\label{eq:ss_discrete1}
	x_{k+1}=A_i x_{k} +B_i u_k
\end{equation}

\subsection{$\mathcal{H}_2$ design}
For controller design, we apply to the switching system~\eqref{eq:ss_discrete1} a $\mathcal{H}_2$ norm minimization approach similar to the one discussed in~\cite{deOliveira99}.
Before defining it, we first derive the controller in the case of regulation, i.e., where $x_k=0$ is the desired steady state and $u_k=Kx_k$ is the control law.
The application of this approach requires the definition of the performance output
\begin{equation}
	z_k = C_p x_k + D_p u_k
\end{equation}
Note that, for LTI systems, in order to make the $\mathcal{H}_2$ approach essentially equivalent to the design of an LQ controller, matrices $C_p$ and $D_p$ must be properly defined as 
$$C_p=\begin{bmatrix}\sqrt{Q}\\0\end{bmatrix},\,D_p=
\begin{bmatrix}0\\\sqrt{R}\end{bmatrix}$$
where $Q$ and $R$ are the weighting matrices commonly used in the LQ control cost.
In this work we minimize the $\mathcal{H}_2$ cost by considering the dynamics of the system in both operating modes, i.e., we solve
\begin{equation}
	\textrm{inf}_{P,K}\textrm{trace}((C_p + D_p K) P (C_p + D_p K)^T)
\end{equation}
subject to the constraint
\begin{equation}
	\label{LMI inequality1}
	(A_i + B_i K) P (A_i + B_i K)^T - P + I < 0
\end{equation}
for both $i={\rm acc}, {\rm dec}$.
The latter is done, first by introducing the new optimization variable $S$ fulfilling
\begin{equation}
	\label{LMI inequality2}
	{S \geqslant (C_p + D_p K) P (C_p + D_p K)^T}
\end{equation}
Then, we define a new gain matrix $L=KP$.
In this way, by resorting to the Schur complement~\cite{BoysLMIs}, the optimization problem above can be formulated as the LMI one
\begin{equation}
	\label{minimization problem}
	{\min_{L,P,S} \textrm{trace} (S)}
\end{equation}
subject to
\begin{equation}
	\begin{bmatrix}
		{S} & {C_p P + D_p L}\\
		(C_p P + D_p L)^T & {P}
	\end{bmatrix} \geqslant 0
\end{equation}
\begin{equation}
	\label{LMI feas problem2}
	\begin{bmatrix}
		{P - A_iP{A}_i^T - A_i L^T {B}_i - B_i L {A}_i^T -I} & {B_i L}\\
		{L^T {B}_i^T} & {P}
	\end{bmatrix}> 0
\end{equation}
for $i={\rm acc}, {\rm dec}$.\\

\subsection{Distance controller design}
\label{subsec:distance_controller}
The control law designed in this work must have the goal of making the robot speed $v$ adapt to the user's speed $v_{\rm VI}$ and, at the same time, to steer the robot curvilinear abscissa $p$ to $p_{\rm VI}+d^o$, i.e., to a distance $d^o$ ahead with respect to the user position $p_{\rm VI}$. Defining the user-robot distance with $d=p-p_{\rm VI}$, since $p-(p_{\rm VI}+d^o)=d-d^o$, the latter goal is equivalent to steer the $d$ to the reference value $d^o$. In light of this discussion, the control law derived in the previous section is modified as follows.
\begin{equation}
\label{eq:dist_ctrl_law}
	u_k = a_{{\rm VI},k}+ K (x_k-x_k^o) 
\end{equation}
where the reference state vector $x_k^o$ is defined as ${x}^o_k = \begin{bmatrix}v_{{\rm VI},k}&p_{\rm VI}+d^o& v_{{\rm VI},k}& a_{{\rm VI},k}\end{bmatrix}^T$ and where $a_{{\rm VI},k}$ denotes the acceleration of the user.
Denoting with $k_j$, $j=1,\dots,4$ the elements of the gain row vector $K$, and according to the control law defined in~\eqref{eq:dist_ctrl_law}, we have to set, at each time instant, $a_{{\rm ref},k}=k_1(v_{{\rm ref},k}-v_{{\rm VI},k})+k_2(d_k-d_k^o)+k_3(v_{k}-v_{{\rm VI},k})+k_4(a_{k}-a_{{\rm VI},k})$.
For practical implementation, we approximate $a_{{\rm ref},k}$, $a_{k}$, and $a_{{\rm VI},k}$ using the implicit Euler discretization method to avoid the regulator equation to display stability issues, and we obtain:
$v_{{\rm ref},k}-v_{{\rm VI},k}=v_{{\rm ref},k-1}-v_{{\rm VI},k-1}+T_s\left(k_1(v_{{\rm ref},k}\right.-v_{{\rm VI},k})+k_2(d_k-d_k^o)+(k_3+\frac{k_4}{T_s})(v_{k}-v_{{\rm VI},k})-\frac{k_4}{T_s}(v_{k-1}-\left.v_{{\rm VI},k-1})\right)$.
In view of this, the real input is, at each time instant, set to
\begin{equation}\begin{array}{ll}v_{{\rm ref},k}&=v_{{\rm VI},k}+\frac{1}{1-k_1T_s}\left(
		(v_{{\rm ref},k-1}-v_{{\rm VI},k-1})\right.\\
		&+k_2T_s(d_k-d_k^o)+(k_3T_s+k_4)(v_{k}-v_{{\rm VI},k})\\
		&-k_4(v_{k-1}-\left.v_{{\rm VI},k-1})\right)\end{array}\label{eq:control1}\end{equation}

\subsection{Integral action}
The control action developed in the previous section may not display good performance in case of uncertainties in the estimation of the user speed $v_{{\rm VI},k}$.
For this reason, the control law can be robustified thanks to the use of a suitable integrator with anti-windup, placed in parallel with respect to the previous controller.
In this way, the control action is computed as $a_{{\rm ref},k}=k_1(v_{{\rm ref},k}-v_{{\rm VI},k})+k_2(d_k-d_k^o)+k_3(v_{k}-v_{{\rm VI},k})+k_4(a_{k}-a_{{\rm VI},k})+k_5 i_k$, where
$$i_{k+1}={\rm sat}_{v_{\rm MAX}/(3 |k_5|)}\left(i_k+T_s(d_k-d_k^o)\right)$$
where $sat_{\bar{x}}(x)=\min(\max(x,-\bar{x}),\bar{x})$. This results in the following control law
\begin{equation}\begin{array}{ll}v_{{\rm ref},k}&=v_{{\rm VI},k}+\frac{1}{1-k_1T_s}\left(
	(v_{{\rm ref},k-1}-v_{{\rm VI},k-1})\right.\\
	&+k_2T_s(d_k-d_k^o)+(k_3T_s+k_4)(v_{k}-v_{{\rm VI},k})\\
	&-k_4(v_{k-1}-\left.v_{{\rm VI},k-1})+k_5T_si_k\right)\end{array}\label{eq:control2}\end{equation}

\subsection{Experimental validation}
\begin{figure}[H]
	\centering
	\includegraphics[width=1\linewidth]{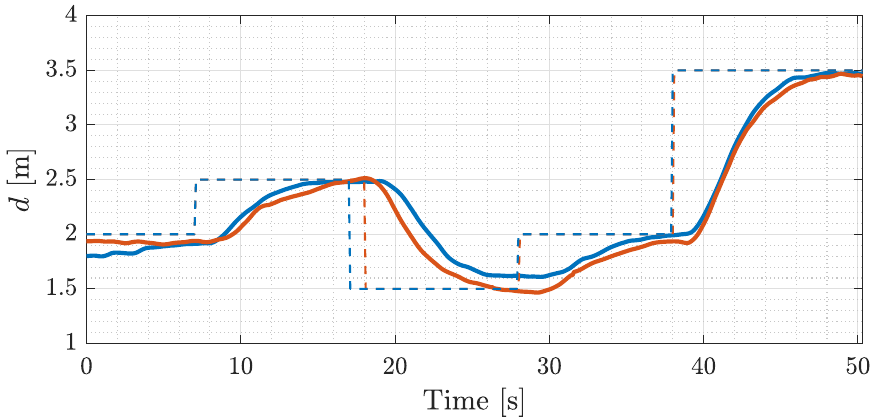}
	\caption{Response (i.e., measured user-robot distance $d$) to step-wise inputs of the distance control systems designed:~\eqref{eq:control1} (blue line) and~\eqref{eq:control2} (red line). Dashed lines: reference distances $d^o$.}
	\label{fig:c_steady1}
\end{figure}

\begin{figure}[H]
	\centering
	\includegraphics[width=1\linewidth]{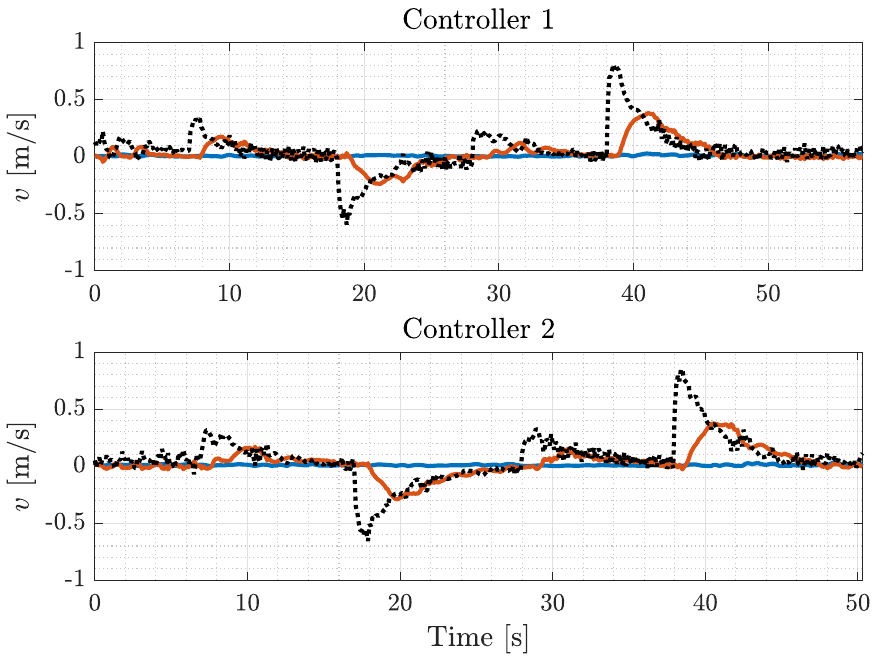}
	\caption{Response to step-wise inputs of the distance control systems~\eqref{eq:control1} (Controller 1), and~\eqref{eq:control2} (Controller 2). For all panels we show: reference Yape velocity $v_{\rm ref}$ (black dashed line); Yape velocity $v$ (red line); user estimated velocity $v_{\rm VI}$ (blue line).}
	\label{fig:c_steady2}
\end{figure}

The controller developed in this section was implemented on the real system and tested.
The tests were done without the use of the Smart Tether System and conducted with the aim of evaluating the performances of the controllers described in this section (Controller 1 and 2 denote~\eqref{eq:control1} and~\eqref{eq:control2}, respectively).
The first tests (see Figures~\ref{fig:c_steady1} and~\ref{fig:c_steady2}) were conducted in case the user was in a fixed position, while the reference distance of Yape changed over time in a step-wise fashion.

As it is possible to notice, e.g., from Figure~\ref{fig:c_steady1}, the $\mathcal{H}_2$-norm minimizing controllers show an overall smooth behavior, quickly reaching the steady state condition.
Controller 2 in particular, endowed with integral action, is able to provide better static performance.

\begin{figure}[H]
	\centering
	\includegraphics[width=1\linewidth]{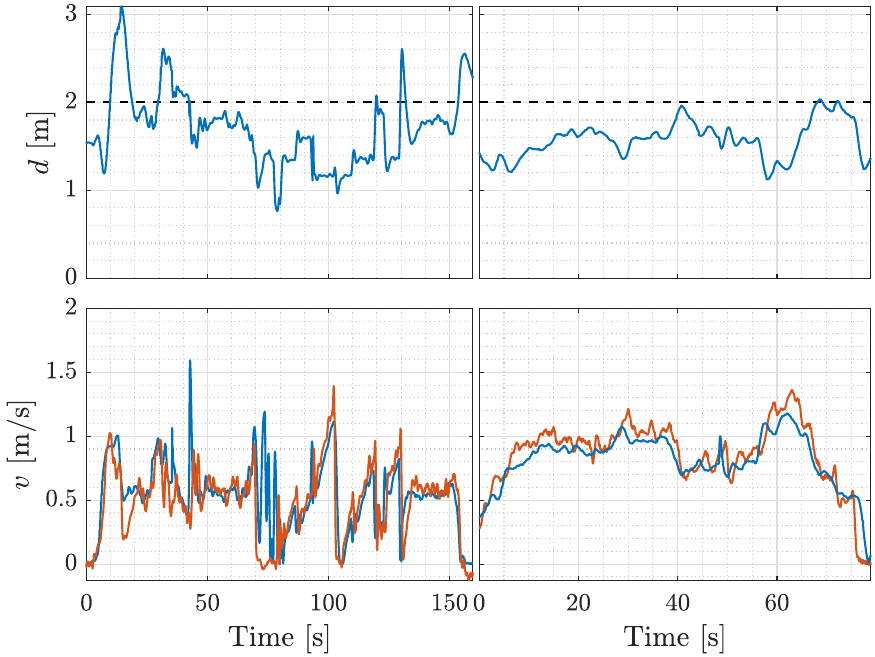}
	\caption{\label{fig:LMI_walking}Response to varying user speed of the distance control systems ~\eqref{eq:control1} (Controller 1, left panel) and~\eqref{eq:control2} (Controller 2, right panel). Top panels: reference distance $d^o$ (black dashed line); measured distance $d$ (blue solid line). Bottom panels: Yape velocity $v$  (red line); user estimated velocity $v_{\rm VI}$ (blue line).}
\end{figure}

\noindent
In the second test, the user moved at a low speed (about 0.5-1 m/s) whereas the reference distance between Yape and the user was kept constant. In Figure~\ref{fig:LMI_walking} the results are displayed. Note that the latter Figure~\ref{fig:LMI_walking} is not reported for the sake of comparison since, as apparent for the bottom panels, the user velocities in the two cases display significantly different dynamical behaviors.
The results displayed in this section witness good both dynamic and static performances of the controllers devised in this work, allowing for the deployment in real-case scenarios like the one described below.
%
%

%
%
%
\subsection{Robot-user distance estimation}
\label{subsec:target_tracking}
The distance control algorithm needs a reliable estimate of the distance and of the velocity of the VI user.
A LiDAR-based target tracking algorithm was implemented to estimate the position and the velocity of the user.
The target tracking is based on~\cite{PARRAVICINI202015440}, which is an Extended Target Tracking algorithm designed and tested on the Yape platform, which is capable of tracking multiple moving objects in the surroundings of the vehicle.
The system is composed of two modules: a target detection and a target tracking one.
The former is tasked with extracting obstacles position measures from the raw point cloud, while the latter aggregates measures belonging to the same objects and estimates their velocities.
While suitable to track incoming vehicles from a distance, the target detection module presented in~\cite{PARRAVICINI202015440} has two major flaws in our application: it compares point clouds within a buffer to extract moving clusters to track and discard static ones.
The downsides with this approach are that (1) it is not possible to track the user while it is standing still and (2) the buffer introduces some delay in the tracking, meaning that also during the initial instants in which the user is moving the robot would have no estimate of its position and velocity.
To address these issues, the target detection module was replaced entirely with one designed specifically for this use-case: a ground removal and clustering algorithm is employed first, with each cluster then accepted for tracking or discarded based on its dimensions.
Furthermore, a track selection process has been devised to estimate which of the tracks obtained from the tracking algorithm is the one relative to the VI user based on the distance from the expected user location.

\begin{figure}[!t]
	\centering
	\includegraphics[width=0.5\linewidth]{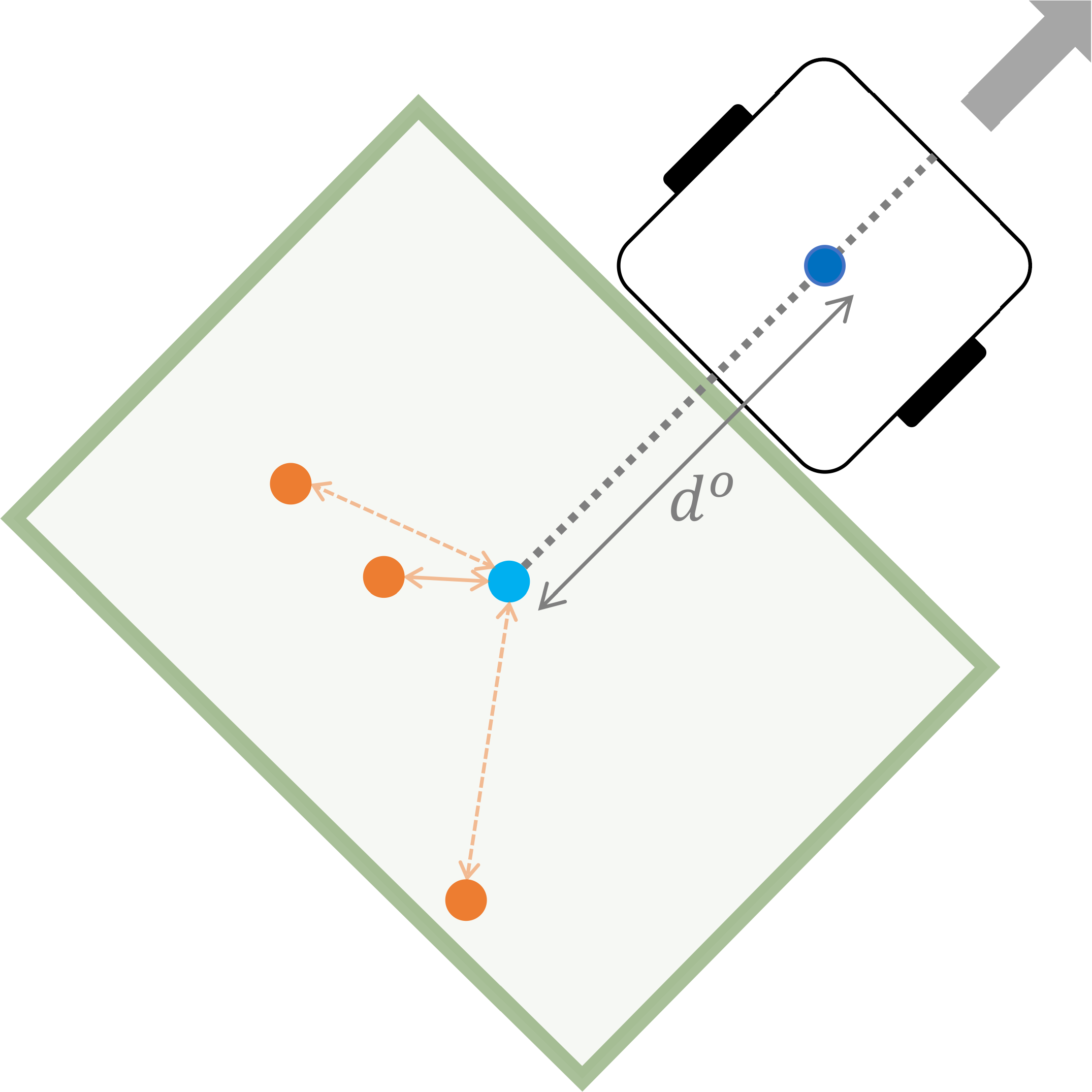}
	\caption{Scheme of the track selection process. The green area is the ROI, the selected track is the one connected with a continuous line to the point behind the robot at the distance reference.}
	\label{fig:distance_control/user_tracking_track_selection}
\end{figure}

First, we run the point cloud through the ground removal and clustering algorithm presented in~\cite{bogoslavskyi_efficient_2017} adapted to account for the variable pitching of the LiDAR sensor.
The result is a point cloud in which no ground points are present and the remaining points are associated with a cluster id. 
Points with the same cluster id are close enough to be considered belonging to the same object.
Note that such clusters include points from nearby pedestrians, the user, vehicles, as well as static obstacles like walls.
Since the objective of the algorithm is to estimate the robot-user distance, we aim to track only clusters with human-like dimensions: we fit oriented bounding boxes with the algorithms implemented in~\cite{rusu_3d_2011} and select for tracking only the clusters with height, width and depth compatible with a human being.
While simple, this approach proved robust enough for our purposes.
To select the track belonging to the user, we employ a set of geometrical rules.
First, we apply a Region Of Interest (ROI) situated on the rear side of the robot, marking all the tracks within the ROI as candidates for selection.
Among them, the selected track is the one closer to the point at distance $d^o$ behind the robot (the point is computed considering the robot's orientation).
The track selection process is represented in Figure \ref{fig:distance_control/user_tracking_track_selection}.

\section{Autonomous Navigation}
\label{sec:autonomous_driving}
\begin{figure}[h]
	\centering
	\includegraphics[width=1\linewidth]{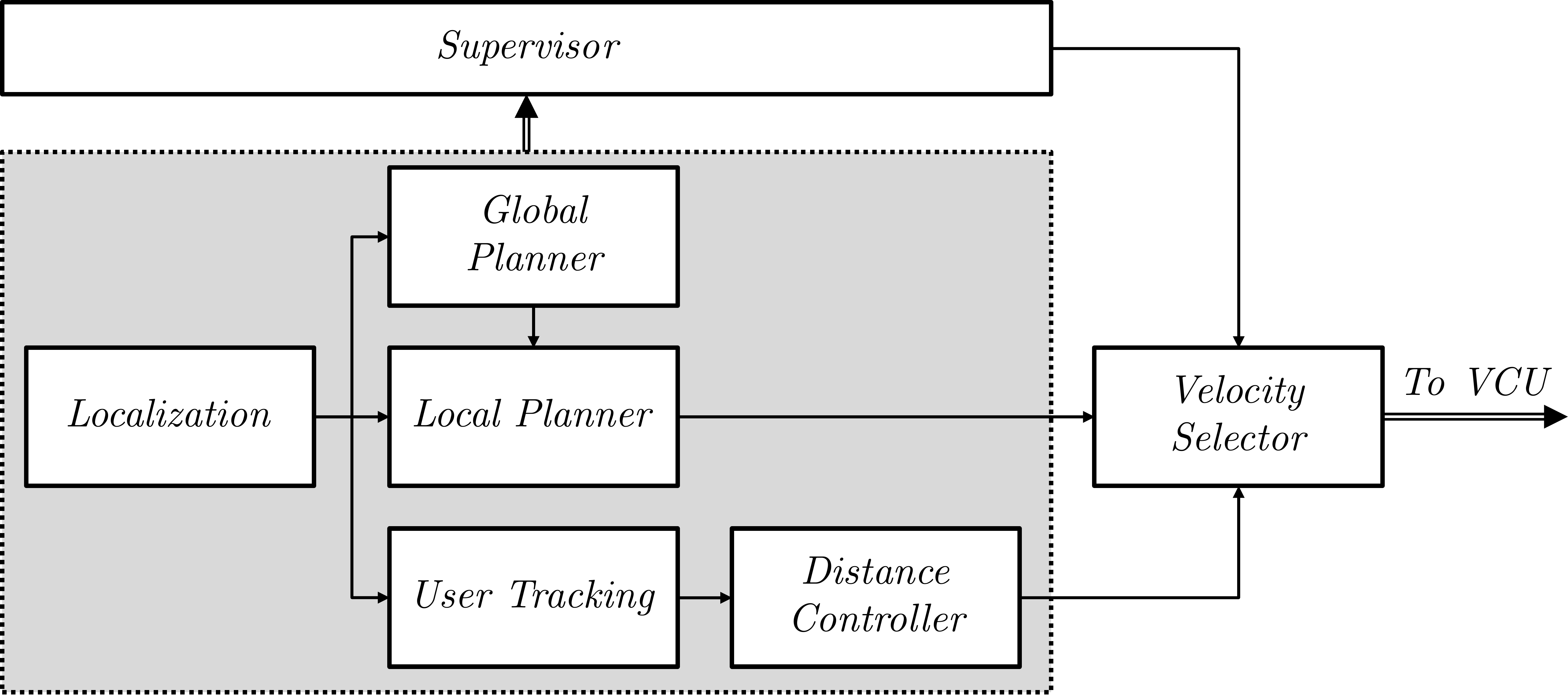}
	\caption{Block scheme of the autonomous driving stack and distance controller.}
	\label{fig:03_autonomous_driving_stack/block_scheme}
\end{figure}

The main components of the autonomous driving stack and their interactions with the distance controller are schematized in Figure \ref{fig:03_autonomous_driving_stack/block_scheme}.

\subsection{Localization}
The localization module is tasked with estimating the vehicle's position and velocity with respect to a world-fixed reference frame.
It employs both GNSS data and a map-based localization algorithm as inputs to a robust two-layered localization architecture.
For more details about the localization architecture the reader is referred to~\cite{mozzarelli_mobile_2024}, whereas the map-based algorithm and the rationale behind its selection are presented in~\cite{hess_real-time_2016,mozzarelli_comparative_2023}, respectively.

\subsection{Global Planner}
The global planner works on a graph structure, selecting the appropriate trajectories connecting the robot's current position with the desired destination.
Since the target environment is complex and subject to rules not easily detectable from on-board sensors (e.g., pedestrian passageways that could be traversed in one direction due to COVID restrictions), the global plans connecting different locations were recorded by teleoperating the robot manually.

\subsection{Local Planner and Collision Avoidance}
As for the local planner, a modified version of the popular Dynamic Window Approach (DWA)~\cite{fox_dynamic_1997} was employed.
DWA is an optimal path planner based on the receding horizon principle.
The robot's position is forward simulated within a prediction horizon under the assumption of constant velocity and yaw rate.
The resulting trajectory is scored via a dedicated cost functions that takes into account of the distance from the global plan, the progress towards the goal, and the clearance from the obstacles.
In our work, based on the modification presented in~\cite{gusmini_development_2020}, we replace the traditional path and goal cost functions with a single one.
Such cost function assigns to each location visited by the robot a cost quadratic in the distance from the global plan and linearly decreasing towards the goal.
Regarding the obstacles cost function, we employed the typical approach of the local costmap, marking LiDAR-measured obstacles on a grid centered around the robot's location.
A trajectory that passes over a marked cell gets discarded to avoid collisions.
Note that the collision check also considers the robot's footprint and some inflation, as to leave some clearance if possible.
The costmap is fed with the output of the ground removal algorithm presented in Section \ref{subsec:target_tracking}.
The robot could encounter obstacles that are challenging to detect from a single point of view and using only LiDAR data, e.g., low-trimmed grass to the side of a concrete sidewalk or negative obstacles like descending steps.
To manage these dangerous conditions, a navigability map obtained using the algorithm proposed in~\cite{mozzarelli_automatic_2023} was inserted as the base layer of the costmap.
This enables the costmap to represent obstacles present in the a-priori available map, as well as the ones detected at runtime by the LiDAR sensor.

\subsection{Velocity Selector}
As was hinted at in Section \ref{subsec:control_system}, both the distance controller and the local planner provide a velocity reference.
During normal operation, i.e., in the absence of obstacles, the robot should track the distance control reference $v^{dist}$.
On the contrary, when an obstacle blocks the robot's movement or forces it to slow down, collision avoidance should have priority, and the robot should slow down or stop according to the velocity setpoint $v^{dwa}$ requested by the local planner.
Therefore, we select as the reference velocity to be sent to the VCU the minimum between the two:
\begin{equation}
	v_{\rm ref} = \min\left\{ v^{dwa}, v^{dist}\right\}
	\label{eq:velocity_selector_v}
\end{equation}
Note that the local planner guarantees that the trajectory generated by the pair $(v^{dwa},\omega^{dwa})$ is collision-free.
The same cannot be stated for the trajectory generated by $(v^{dist},\omega^{dwa})$.
To maintain the collision-free guarantees, the yaw rate reference is scaled as well, as to obtain an arc with the same curvature as the one selected by the local planner:
\begin{equation}
  \omega_{\rm ref} = \left\lvert\frac{v^{dist}}{v^{dwa}}\right\rvert  \omega^{dwa}
	\label{eq:velocity_selector_w}
\end{equation}
Finally, the distance controller velocity is saturated to be positive and smaller or equal to the maximum velocity selectable by the local planner.
\subsection{Supervisor}
\begin{figure}[h]
	\centering
	\includegraphics[width=0.85\linewidth]{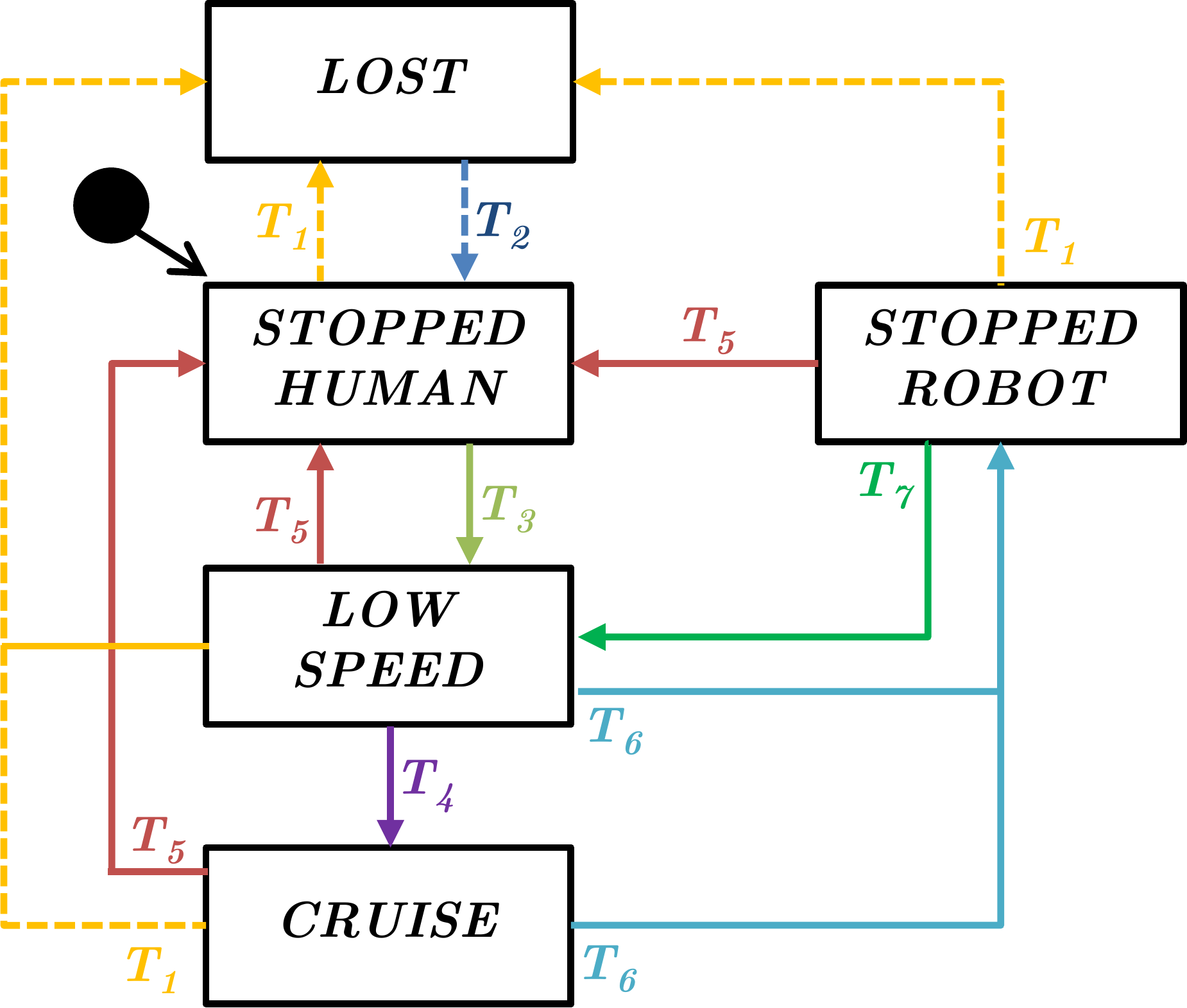}
	\caption{Finite State Machine of the supervisor.}
	\label{fig:03_autonomous_driving_stack/behavior_manager}
\end{figure}

A supervisory Finite State Machine (FSM), depicted in Figure \ref{fig:03_autonomous_driving_stack/behavior_manager}, was designed to guarantee the safety of the device and handle possible interactions between modules.
Depending on the state of the FSM, the velocity selector can override the default behavior in \eqref{eq:velocity_selector_v} and \eqref{eq:velocity_selector_w}.
In the first three states ($LOST$, $STOPPED~HUMAN$ and $STOPPED~ROBOT$) the velocity selector commands the robot to stand still.
Transitions $T_1$ and $T_2$ bring in and out of the $LOST$ state and are triggered based on the quality of the localization solution.
The $STOPPED~HUMAN$ state is activated ($T_5$) when there is a user-related reason for stopping: either the user track is lost, the user is too far, a manual stop has been requested or the desired destination has been reached.
To exit from such a state and start moving again only with manual acknowledgment by the user ($T_3$).
State $STOPPED~ROBOT$ is entered with transition $T_6$ when the robot requests a stop, typically because a blocking obstacle was detected.
Movement from this state ($T_7$) is possible without manual acknowledgment, as soon as the robot finds a viable local plan.
The $CRUISE$ state is active during nominal travel conditions, and in this case the velocity selector applies the equations described above \eqref{eq:velocity_selector_v} \eqref{eq:velocity_selector_w}.
State $LOW~SPEED$, instead, was inserted to handle the starting phases of the navigation, as treating them nominally caused the user to collide with the robot before it had the chance to pick up speed and start moving.
In this state the velocity selector forwards the local planner velocities directly, bypassing the distance controller and guaranteeing a swifter response.
This solution is deemed safe as the robot is free to travel at the maximum speed for a few instants only and after acknowledgment by the user.
As soon as the robot and the user pick up speed, transition $T_4$ gets triggered, enabling the distance controller.

\section{Experimental campaign: wayfinding in healthcare facilities}
\label{sec:validation}
\begin{figure}[h]
	\centering
	\includegraphics[width=1\linewidth]{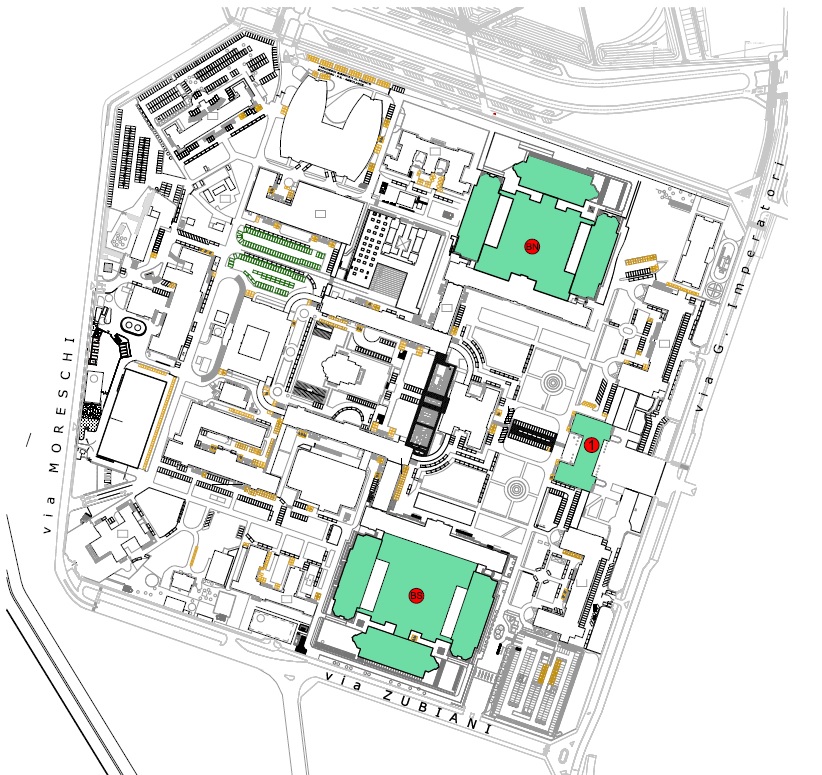}
	\caption{\label{fig:demo_map0} ASST Grande Ospedale Metropolitano Niguarda map. The main entrance and the Blocco Sud and Blocco Nord buildings are depicted in green.}
\end{figure}

\begin{figure}[H]
	\centering
	\includegraphics[width=\linewidth]{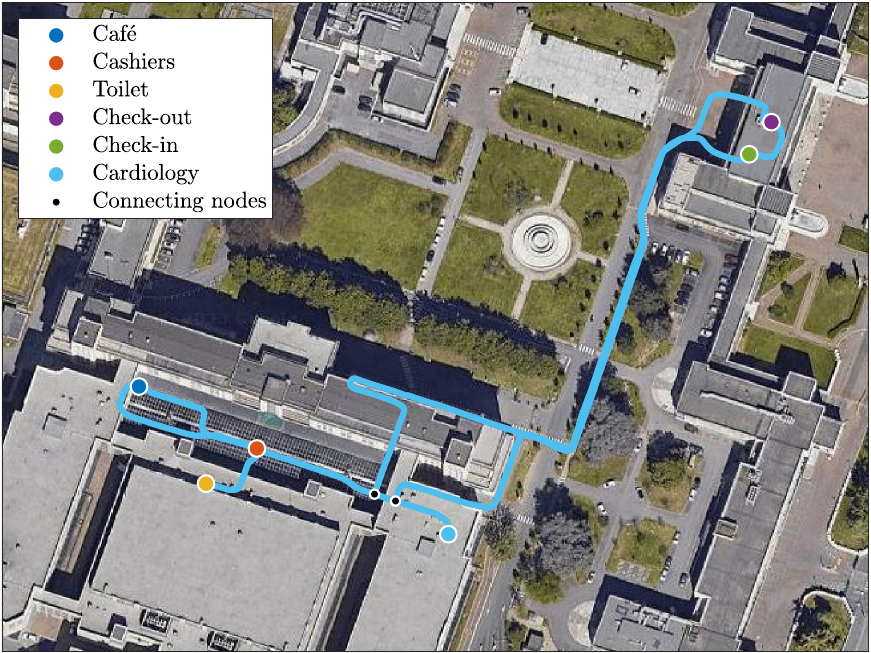}
	\caption{Global plans}
	\label{fig:demo/niguarda_global_plans}
\end{figure}

\noindent
The BUDD-e project involved an experimentation hosted at the ASST Grande Ospedale Metropolitano Niguarda, whose map is displayed in Figure \ref{fig:demo_map0}.
During the demo, BUDD-e had to guide visually impaired volunteers from the hospital's entrance (Padiglione 1) to the cardiology department (Blocco Sud) including optional deviations, e.g., to the coffee bar.
The selected path has been scanned with Yape's LiDAR sensor, and the track has been established and recorded, see Figure~\ref{fig:demo/niguarda_global_plans}.
The experimental campaign consisted of a number of preliminary user-centered tests, whose results are shown in Section~\ref{subsec:preliminary_tests}, and a final demo described in Section~\ref{subsec:final_demo}.
Preliminary trials have been conducted with and without the user's presence, with the aim of evaluating BUDD-e's ability to autonomously navigate the track and avoid obstacles.\smallskip\\
The Politecnico di Milano Ethics Committee approved all the activities related to the project (n. 23/2022, 06/06/2022), which were performed according to the ethical principles set out in the Declaration of Helsinki. All participants were informed about data collection purposes and gave written informed consent prior to their participation. 

\subsection{User-centred tests}
\label{subsec:preliminary_tests}
\noindent
First, a campaign of tests was conducted with the participation of 10 volunteers. Basic information about the participants is reported in Table \ref{tab:Participants}.
\begin{table}[h]
	\centering
	\begin{tabular}{ | l | p{1.5cm} | p{1cm} | p{2.5cm} |}
		\hline
		& \textbf{Gender} & \textbf{Age} & \textbf{Impairment} \\
		\hline
		Volunteer 1 & Female & 57 & Totally blind \\ \hline
		Volunteer 2 & Male & 53 & Totally blind\\ \hline
		Volunteer 3 & Male & 38 & Totally blind \\ \hline
		Volunteer 4 & Male & 61 & Totally blind \\ \hline
		Volunteer 5 & Male & 68 & Totally blind\\ \hline
		Volunteer 6 & Male & 55 & Visually impaired\\ \hline
		Volunteer 7 & Male & 60 & Totally blind\\ \hline
		Volunteer 8 & Female & 52 & Totally blind\\ \hline
		Volunteer 9 & Male & 53 & Visually impaired\\ \hline
		Volunteer 10 & Male & 22 & Visually impaired\\ \hline
		\end{tabular}
	\caption{Volunteers' information (user-centered tests)}
	\label{tab:Participants}
\end{table}
These volunteers performed preliminary tests on the functionalities of BUDD-e, providing valuable insights and feedback.
Based on their inputs, adjustments were adaptively made to enhance the robot's performance.
These modifications included parameter calibrations and reference distance refinements.\\
After their experience with BUDD-e, volunteers were asked to provide their answers to a series of questions, compiling two different questionnaires.
%
\subsubsection{System Usability Scale (SUS) questionnaire}
\label{sec:Questionnaire}
\noindent
The first questionnaire was designed according to the System Usability Scale (SUS)~\cite{SUS}, a tool that allows to quantify the usability of a device.
SUS consists of a 10-item questionnaire featuring five response options, within a range from 'Strongly Disagree' (1) to 'Strongly Agree' (5).
The collected answers were organized and presented in Figure \ref{fig:SUS}.
The SUS scores were calculated for each user and are reported in Table \ref{tab:SUS}.
The latter are computed as follows: for even-numbered questions, we subtracted 1 from the response; for odd-numbered questions, we subtracted the response from 5; the scores obtained in each question are added together; the final result is multiplied by 2.5.
\begin{table}[h]
	\centering
	\begin{tabular}{|c|c|}
		\hline
		Volunteer 1 & 87,5\\
		Volunteer 2 & 100\\
		Volunteer 3 & 52,5\\
		Volunteer 4 & 100\\
		Volunteer 5 & 87,5\\
		Volunteer 6 & 82,5\\
		Volunteer 7 & 100\\
		Volunteer 8 & 87,5\\
		Volunteer 9 & 80\\
		Volunteer 10 & 80\\
		\hline
		Average & 85,75\\
		\hline
	\end{tabular}
	\caption{Usability index}
	\label{tab:SUS}
\end{table}
The following ranges can be used to interpret the SUS results:
\begin{itemize}
	\item 0-25: worst immaginable
	\item 25-35: poor
	\item 35-55: acceptable
	\item 55-75: good
	\item 75-100: best immaginable
\end{itemize}
BUDD-e received an average usability score of 85.75 which, according to the scale presented above, is highly satisfactory.

\subsubsection{Technology Acceptable Model (TAM) questionnaire}
\label{sec:Questionnaire}
\begin{figure*}[t!]
	\centering
  \subfloat[]{
	\includegraphics[width=.475\linewidth]{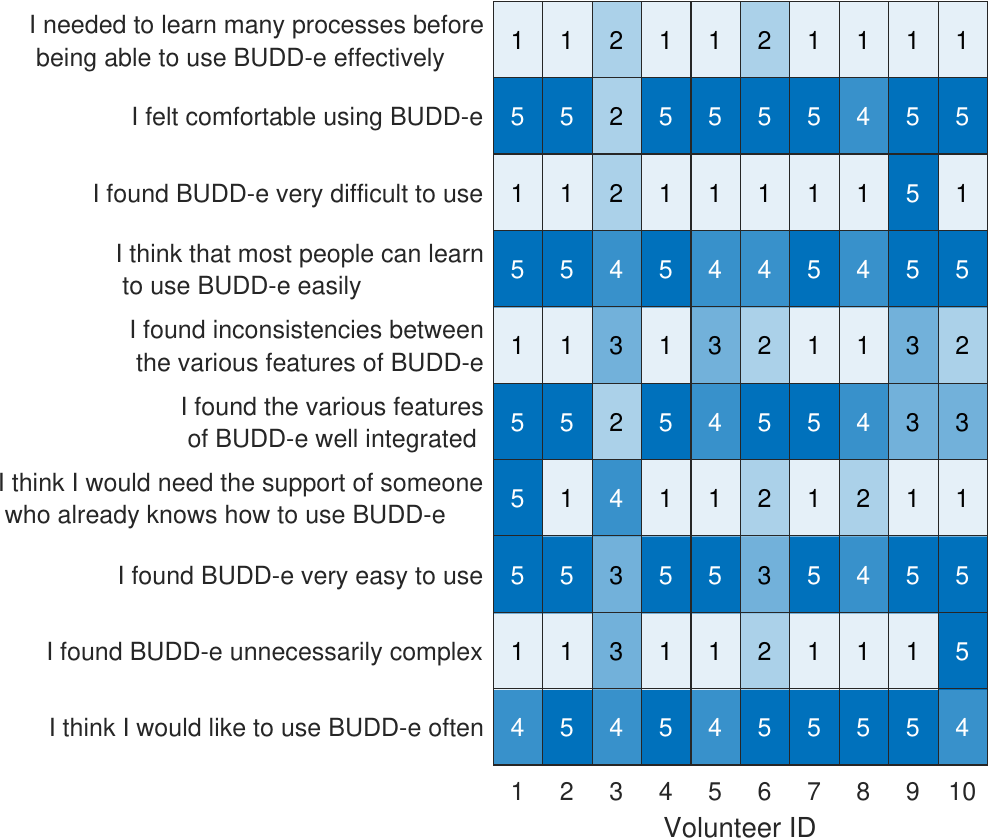}\label{fig:SUS}}
	\hfill
  \subfloat[]{
	\includegraphics[width=.475\linewidth]{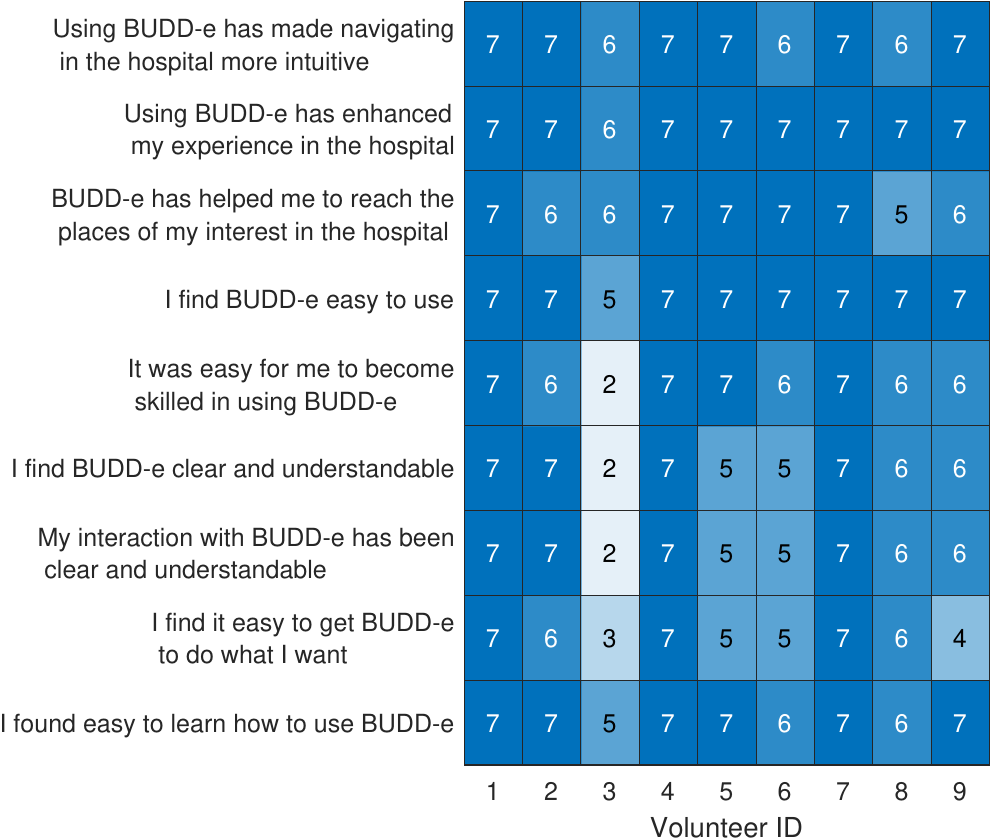}\label{fig:TAM}}
	\caption{System Usability Scale \protect\subref{fig:SUS} and Technology Acceptance Model \protect\subref{fig:TAM} questionnaire results.}
\end{figure*}

\noindent
The second questionnaire has been designed according to the Technology Acceptance Model (TAM)~\cite{TAM} to analyze how users are prone to accept and use the new technology. 
The TAM consists of a 9-item questionnaire (in the following Q$i$ is used to denote the $i$-th question) featuring seven response options, on a scale from 'Extremely unlikely' (1) to 'Extremely likely' (7).
In particular, the model studies the factors that influence the user's feeling about a new technology, namely perceived usefulness (PU, Q1-Q3) and perceived ease-of-use (PEOU, Q4-Q9).
The collected answers have been organized and presented in Figure \ref{fig:TAM}.
Only nine out of ten volunteers answered.
The results are the following. Q1: $6.6667\pm0.4714$; Q2: $6.8889\pm0.31427$; Q3: $6.4444\pm0.68493$; Q4: $6.7778\pm0.62854$; Q5: $6\pm1.4907$; Q6: $5.7778\pm1.5476$; Q7: $5.7778\pm1.5476$; Q8: $5.5556\pm1.3426$; Q9: $6.5556\pm0.68493$.
Overall, the PU-related and PEOU-related questions obtained an evaluation of $6.6667\pm0.54433$ and $6.0741\pm1.3451$, respectively.
Both PU and PEOU indexes are definitely positive but, while the perceived usefulness of BUDD-e received an extremely high score, there is some room for improvement as far as the ease-of-use is regarded.

\subsubsection{Conclusions about the user-centered tests}
\label{sec:user tests conclusions}
\noindent
The results obtained from the two questionnaires exhibit relatively high scores in general, although some volunteers had some difficulties during BUDD-e trials.
Volunteer 3, for example, stated that he did not feel secure, primarily due to technical issues that arose with the Smart Tether System during the test.
This problem made him not get to understand the functioning of BUDD-e and he didn't feel comfortable during its use, as stated in his surveys' responses.\\
The two questionnaires reveal a further issue, noticed during the trials.
All users with a visual residual (i.e., Volunteer 6, Volunteer 9, and Volunteer 10), stated that there are some inconsistencies and poorly integrated features in BUDD-e.
This, in our opinion, is partially due to the fact that, because of their partial sight, they did not properly use the functionalities of the robot and, in some cases, got in conflict with the system.
For instance, since they can see BUDD-e, they tend not to rely on the force provided by the Smart Tether System for their motion and they \quotes{anticipate} the instructions provided by it.
\subsection{Final demonstration}
\label{subsec:final_demo}
\begin{figure*}[h]
	\centering
	\begin{minipage}{.32\linewidth}
		\includegraphics[width=\textwidth]{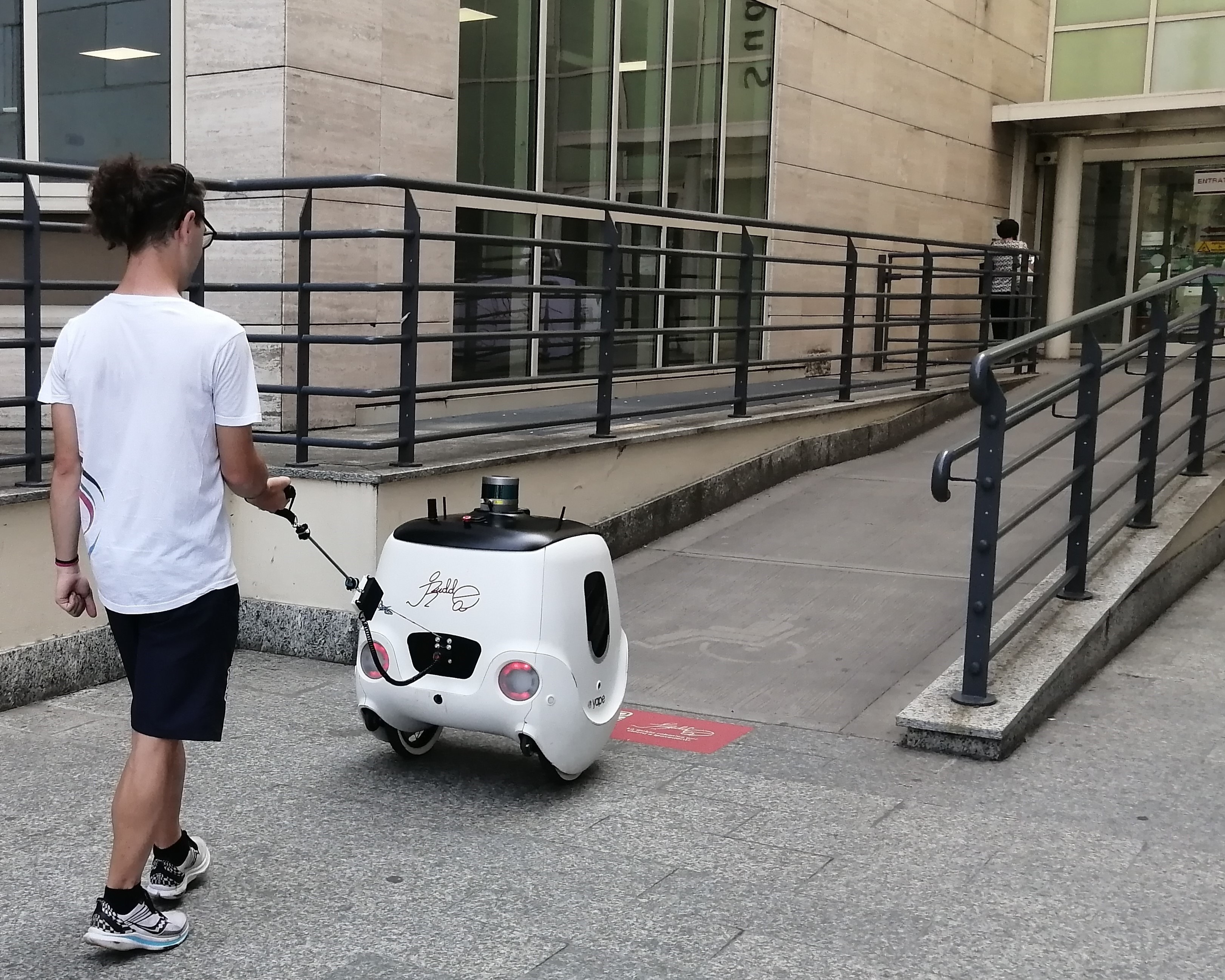}
	\end{minipage}
	\hfill
	\begin{minipage}{.32\linewidth}
		\includegraphics[width=\textwidth]{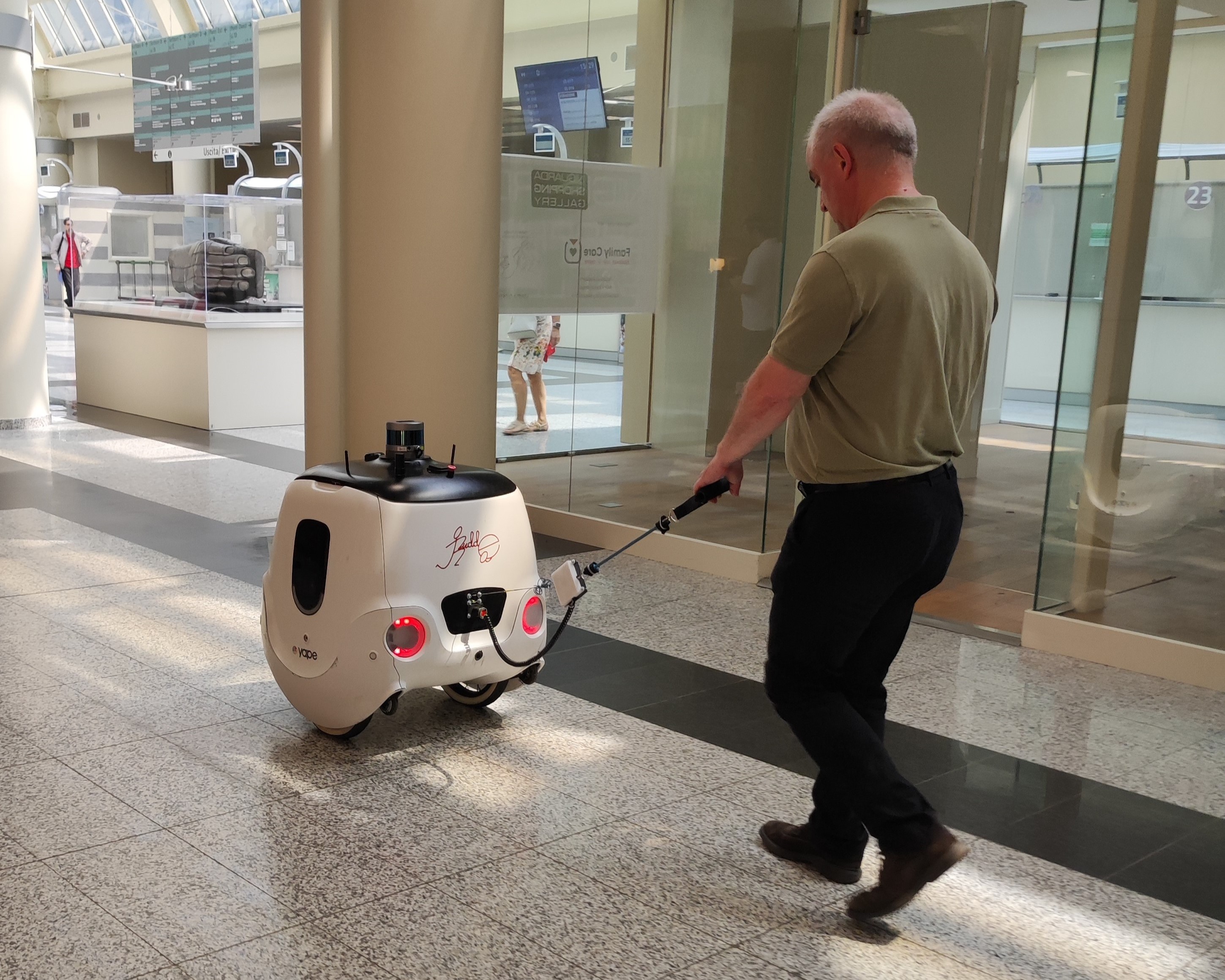}
	\end{minipage}
	\hfill
	\begin{minipage}{.32\linewidth}
		\includegraphics[width=\textwidth]{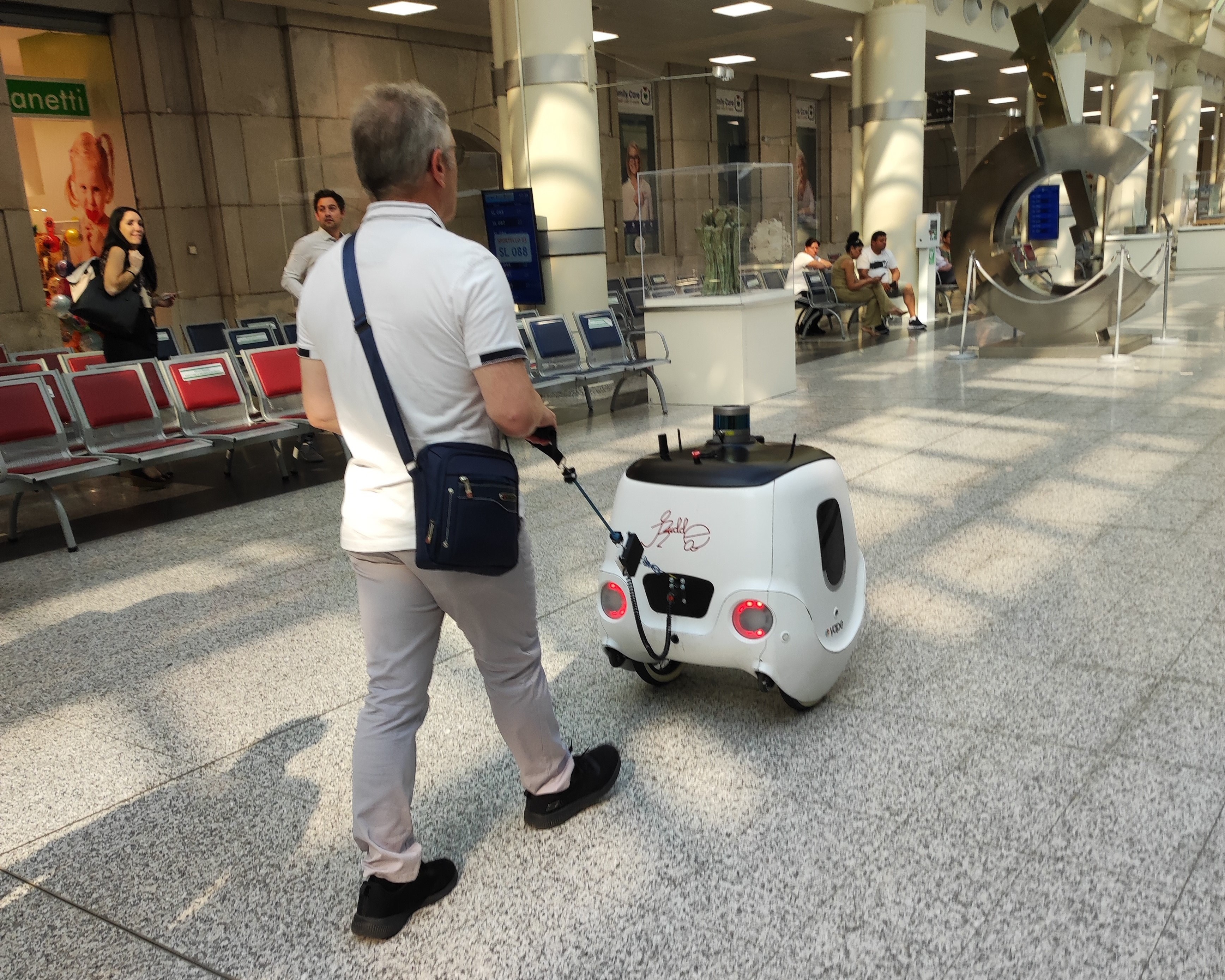}
	\end{minipage}
	\caption{Photos of the experimental campaign within the hospital.}
	\label{fig:users_photos}
\end{figure*}

\noindent
The final demonstration, taking place on 27th June 2023, hosted five visually impaired volunteers, each covering a complete route of the  ASST Grande Ospedale Metropolitano Niguarda selected path.
Information about the participants is reported in Table \ref{tab:Participants2}, whereas the pictures of the tests are displayed in Figure~\ref{fig:users_photos}.
\begin{table}[H]
	\centering
	\begin{tabular}{ | l | p{1.5cm} | p{1.5cm} | p{2.5cm} |}
		\hline
		& \textbf{Gender} & \textbf{Age} & \textbf{Impairment} \\
		\hline
		Volunteer 1 & Male & 53 & Totally blind\\ \hline
		Volunteer 2 & Male & 22 & Visually impaired\\ \hline
		Volunteer 3 & Male & 55 & Visually impaired\\ \hline
		Volunteer 4 & Male & 68 & Totally blind\\ \hline
		Volunteer 5 & Male & 61 & Totally blind \\ \hline
	\end{tabular}
	\caption{Volunteers' information (final demo)}
	\label{tab:Participants2}
\end{table}
Before the final demonstration, a loudspeaker was installed on BUDD-e to provide simple vocal indications.
Auditory feedbacks were triggered when BUDD-e started its route and when the final destinations (cardiology department, entrance, or bar) were reached.
Note that no structured questionnaires were conducted in this phase.
In the initial phase, the volunteers were positioned behind BUDD-e.
The starting phase was acknowledged by supervision personnel through the usage of a joystick.
The goal, for future iterations, is to integrate a button on the handle for the starting procedure as well as voice commands to select the destination and other more advanced interactions.

Due to the redundancy of information and the similarity among the tests, in this section we summarize only their most noteworthy characteristics and other pertinent details.

\subsubsection{Autonomous Navigation}
\begin{figure*}
  \centering
  \subfloat[]{
    \includegraphics[width=0.24\textwidth]{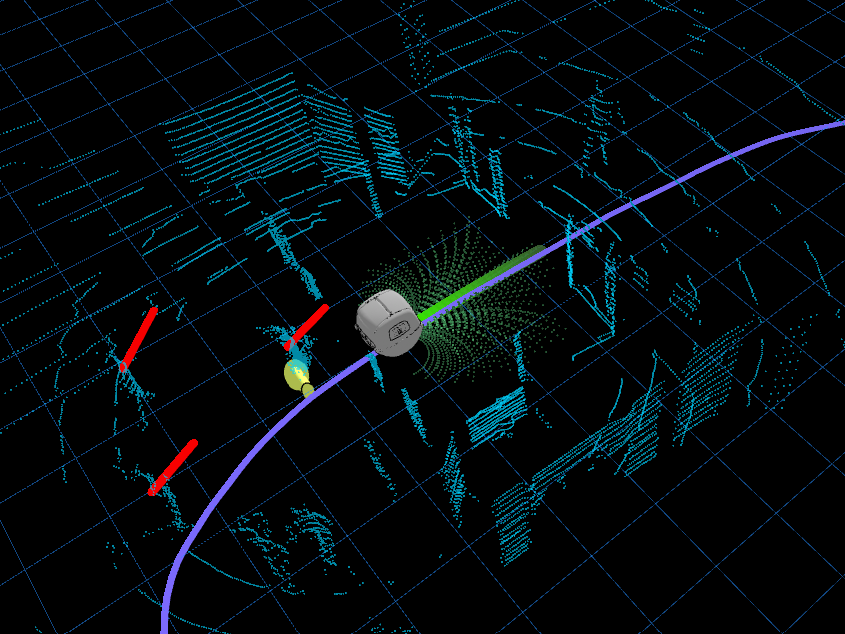}%
  \label{fig:niguarda_validation/foxglove_entrance_doors}}%
  \hfill%
  \subfloat[]{
    \includegraphics[width=0.24\textwidth]{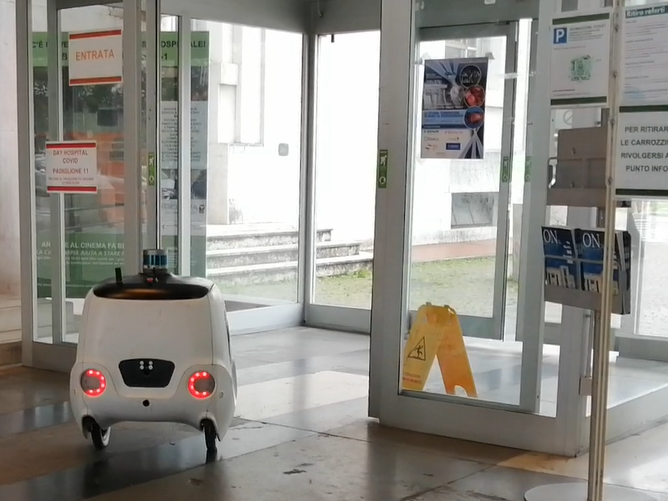}%
  \label{fig:niguarda_validation/photo_entrance_doors_no_user}}%
	\hfill%
  \subfloat[]{\includegraphics[width=0.24\textwidth]{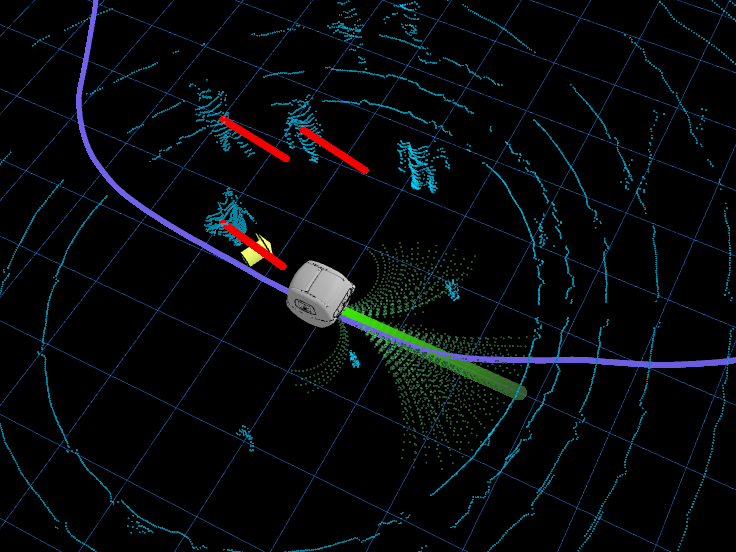}%
  \label{fig:niguarda_validation/foxglove_entrance_bollards}}%
  \hfill%
  \subfloat[]{\includegraphics[width=0.24\textwidth]{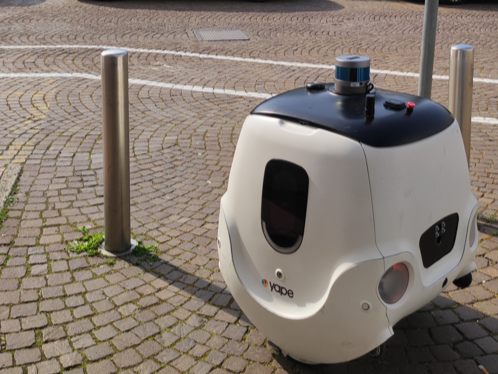}%
  \label{fig:niguarda_validation/photo_entrance_bollards_no_user}}%
  \caption{Snapshots of autonomous navigation: exiting from building 1 through double automatic doors \protect\subref{fig:niguarda_validation/foxglove_entrance_doors}\protect\subref{fig:niguarda_validation/photo_entrance_doors_no_user} and passing through the bollards \protect\subref{fig:niguarda_validation/foxglove_entrance_bollards}\protect\subref{fig:niguarda_validation/photo_entrance_bollards_no_user}. Pictures and screenshots are referred to the same location but not the same experiment.}
  \label{fig:niguarda_validation/foxglove_photos_entrance}
\end{figure*}

\begin{figure*}
  \centering
  \subfloat[]{
    \includegraphics[width=0.245\textwidth,valign=m]{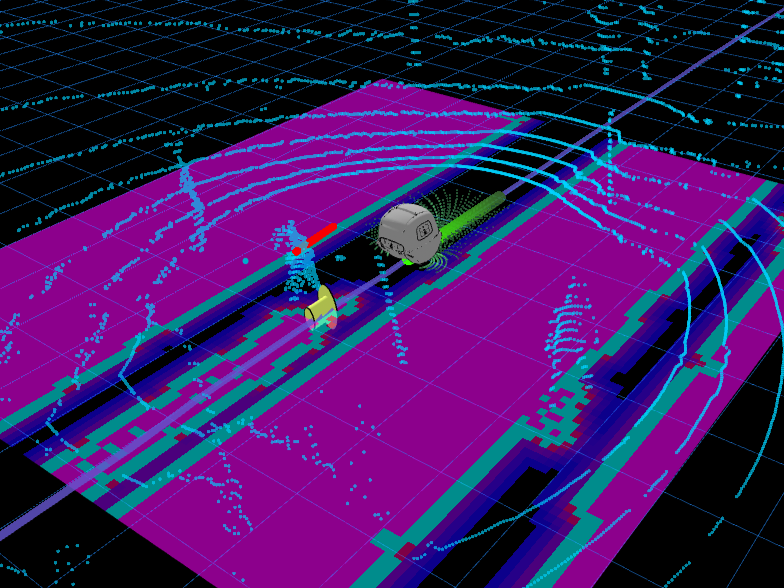}%
  \label{fig:niguarda_validation/foxglove_sidewalk_grass}}%
  \hfill%
  \subfloat[]{
    \includegraphics[width=0.24\textwidth,valign=m]{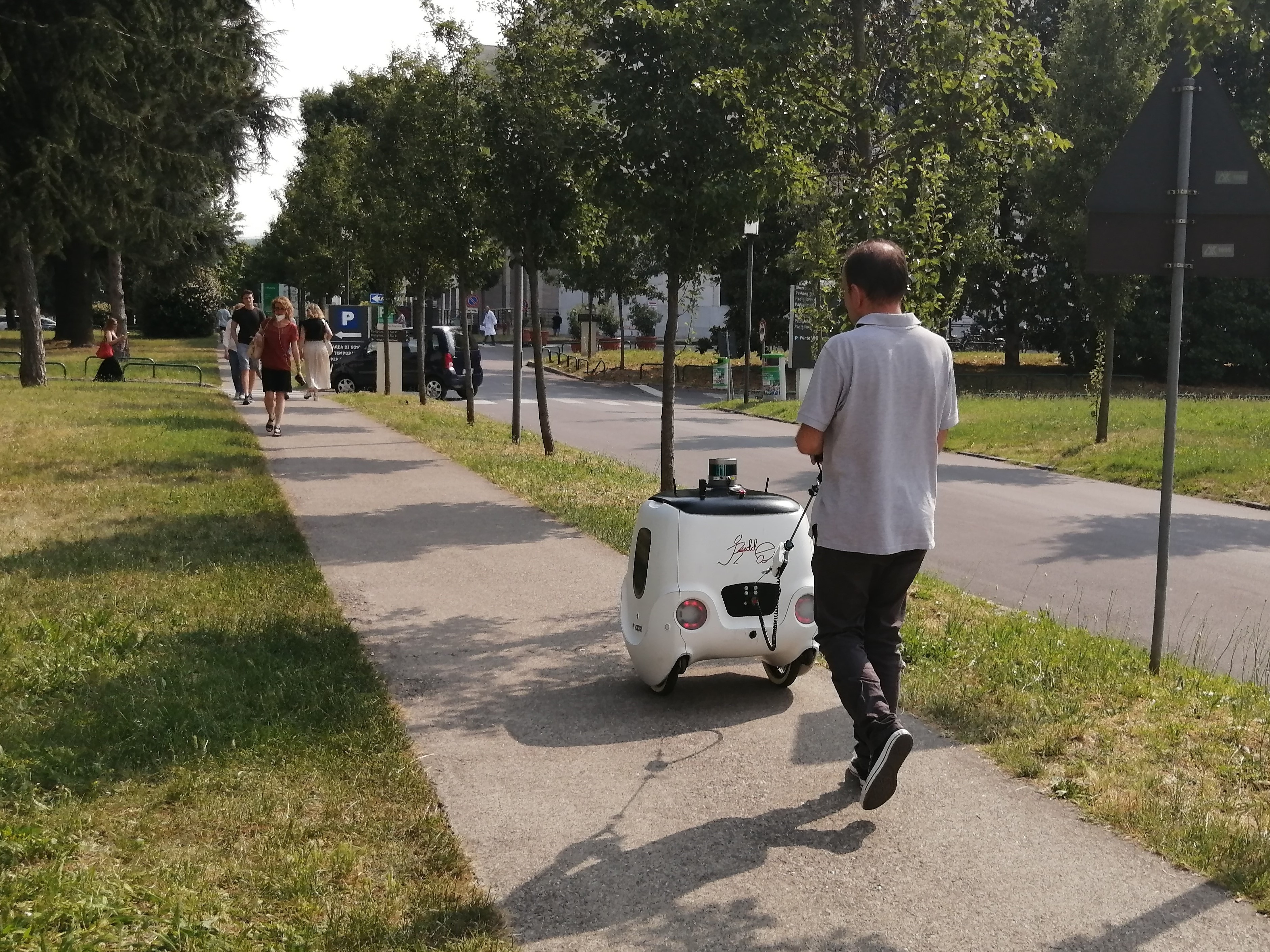}%
  \label{fig:niguarda_validation/photo_sidewalk_grass_user}}%
	\hfill%
	\subfloat[]{
    \includegraphics[width=0.245\textwidth,valign=m]{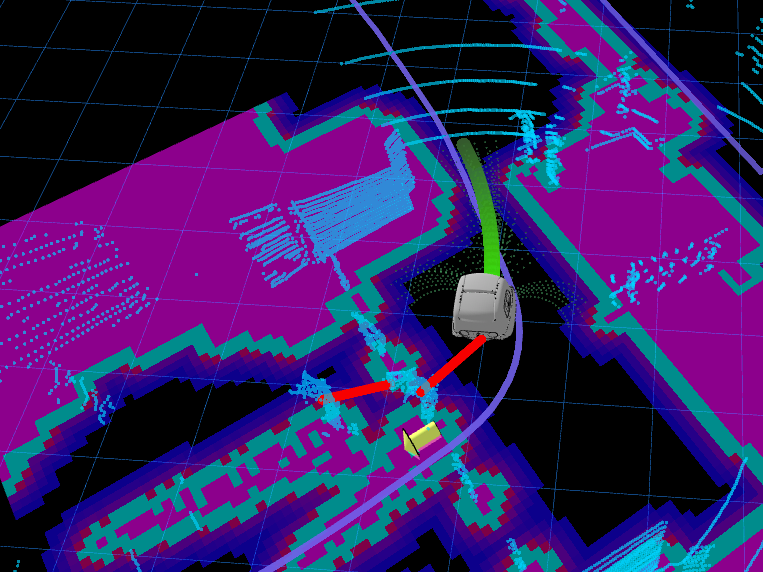}%
  \label{fig:niguarda_validation/foxglove_blocco_sud_stairs}}
  \hfill%
	\subfloat[]{
    \includegraphics[width=0.24\textwidth,valign=m]{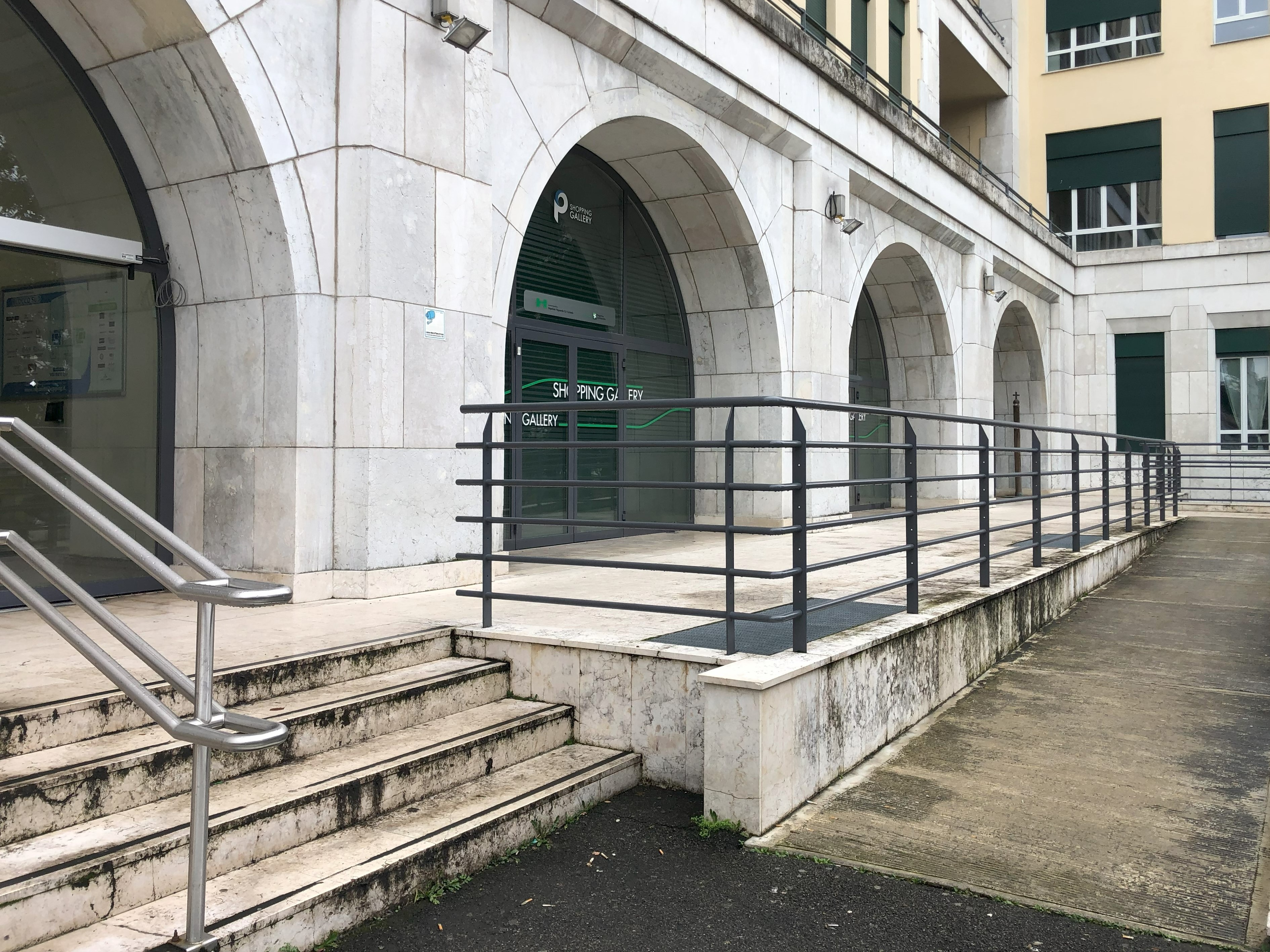}%
  \label{fig:niguarda_validation/photo_blocco_sud_stairs_no_user}}%
  \caption{Snapshots of autonomous navigation: moving from the entrance building to \textit{Blocco Sud} \protect\subref{fig:niguarda_validation/foxglove_sidewalk_grass}\protect\subref{fig:niguarda_validation/photo_sidewalk_grass_user} and exiting the \textit{Blocco Sud} \protect\subref{fig:niguarda_validation/foxglove_blocco_sud_stairs}\protect\subref{fig:niguarda_validation/photo_blocco_sud_stairs_no_user}. Pictures and screenshots are referred to the same location but not the same experiment. \label{fig:niguarda_validation/foxglove_photos_blocco_sud_stairs}}
\end{figure*}

\begin{figure}[!h]
  \centering
  \includegraphics[width=.9\linewidth]{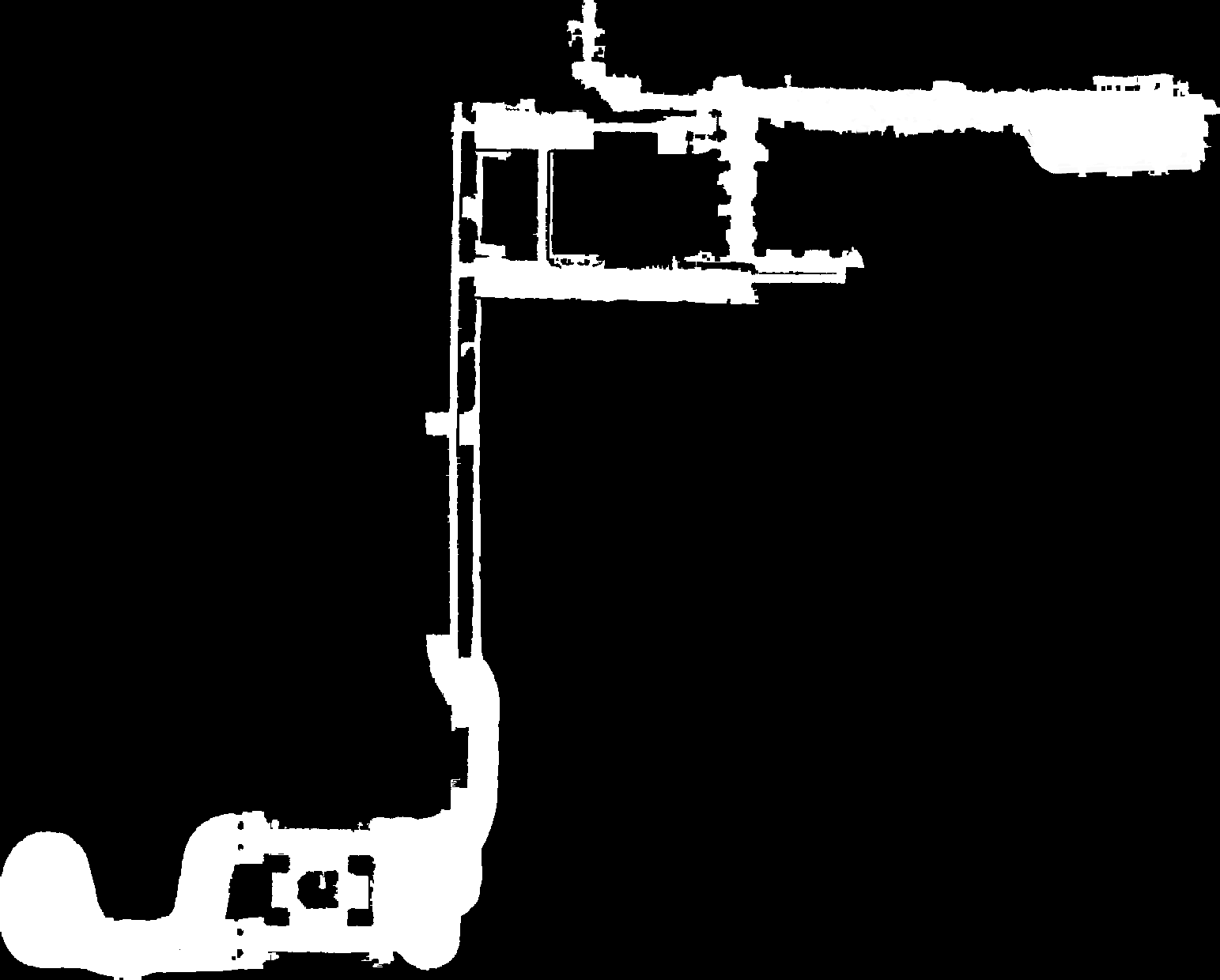}
  \caption{Navigability map of the Niguarda hospital. Black areas are considered as not navigable, white areas as safe for traversal.}
  \label{fig:niguarda_validation/navigability_map}
\end{figure}

In this paragraph we provide some remarks on the autonomous navigation performence of the robotic guide, highlighting the effectiveness of the implemented solution.
Figure \ref{fig:niguarda_validation/foxglove_entrance_doors} displays some snapshots of the starting phase of the navigation, when BUDD-e was exiting from the entrance building through two automatic doors (Figure \ref{fig:niguarda_validation/foxglove_entrance_doors}) and successfully navigating through the tightly spaced bollards (Figure \ref{fig:niguarda_validation/foxglove_entrance_bollards}).
In the screenshots, the blue line represents the global plan, the green line the local plan, and the green dotted lines are the feasible plans evaluated by the DWA.
The user is identified by a yellow marker, while human tracks are represented by red arrows, indicating their velocities.
Figures \ref{fig:niguarda_validation/foxglove_sidewalk_grass} and \ref{fig:niguarda_validation/photo_sidewalk_grass_user} show the navigation in the grass-edged sidewalk.
Notice how the local costmap (purple bands) represents the sidewalk edges even if they are not direct LiDAR hits.
This is the effect of including the navigability map (represented in Figure~\ref{fig:niguarda_validation/navigability_map}) in the costmap.

Entering the \textit{Blocco Sud}, as well as the indoor navigation is quite straightforward, with the main challenge being the large crowd present in the cashiers' desks area. 
Note that this outdoor-indoor switch highlights the versatility of the employed localization algorithm.
Finally, Figures~\ref{fig:niguarda_validation/foxglove_blocco_sud_stairs} and \ref{fig:niguarda_validation/foxglove_photos_blocco_sud_stairs}d show the most challenging phase of the navigation: the exit from \textit{Blocco Sud}.
Notice how, thanks to the inclusion of the navigability map, the negative obstacle represented by the stairs is present in the local costmap (purple areas), even if they are not present in the LiDAR scan.

\subsubsection{User tracking}
\begin{figure}[!h]
  \centering
	\includegraphics[width=0.9\linewidth]{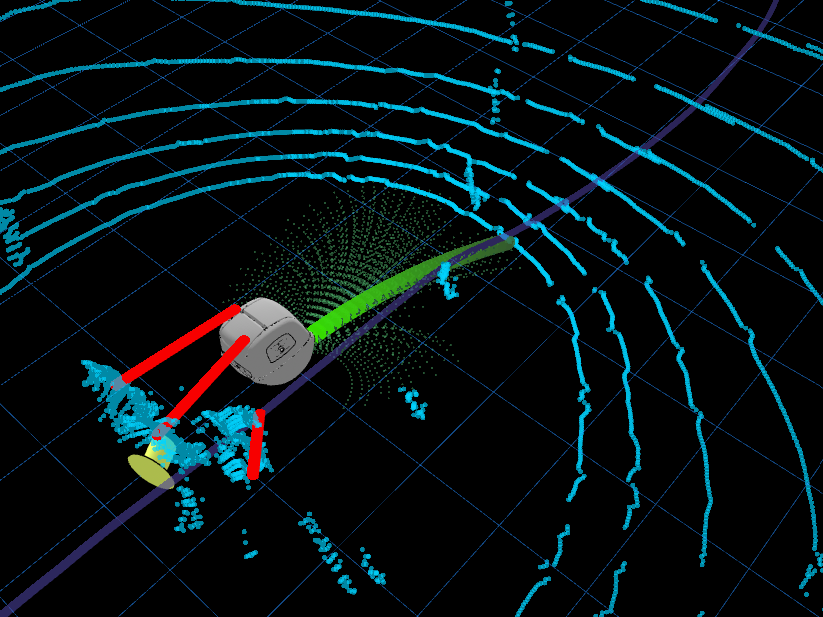}%
  \caption{Example of user tracking. Target tracking tracks are represented in red (the length and direction of the line represent the direction and magnitude of the pedestrian speed). The selected track is indicated by the yellow marker.}
  \label{fig:niguarda_validation/foxglove_user_tracking}
\end{figure}

Figure \ref{fig:niguarda_validation/foxglove_user_tracking} shows snapshots illustrating the effectiveness of the target tracking and track selection process in two challenging conditions.
In the presented scenario two people walk side-by-side with the actual user.
This is a potentially critical situation, given that three persons are close to the supposed position of the user and the distance controller could end up tracking the wrong person.
The red arrows represent the humans' velocities, indicating that all three people are correctly tracked.
The selected track is correctly indicated by the yellow marker.

\subsubsection{User-robot interaction}
\begin{figure}[!h]
  \centering
	\includegraphics[width=0.9\linewidth]{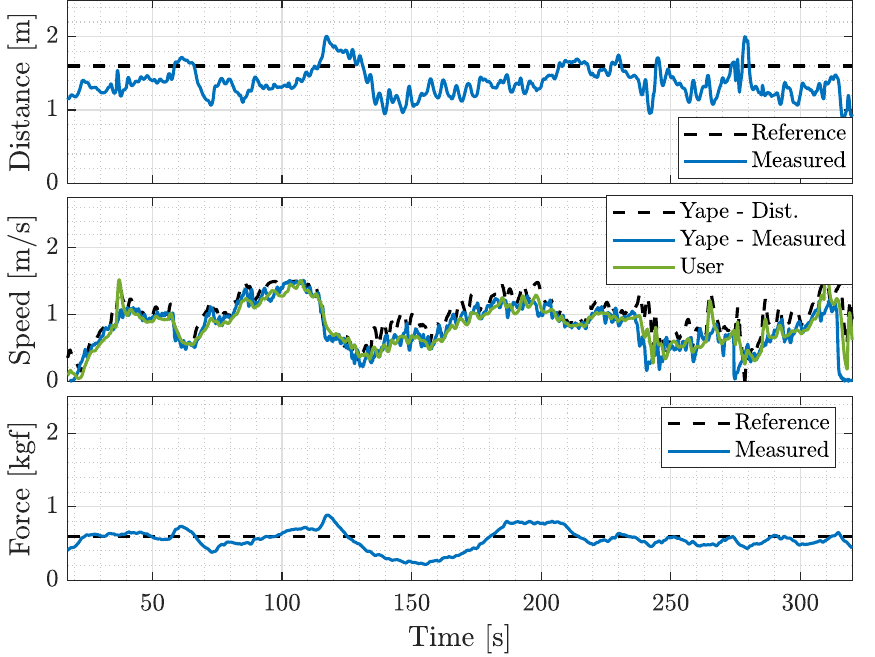}%
  \caption{Distance, velocity and tether force during one of the experiments starting at the entrance building and ending at the cardiology ward.}
  \label{fig:niguarda_validation/entrance_to_cardio_distance_ctrl}
\end{figure}

\begin{figure}[!h]
  \centering
	\includegraphics[width=0.9\linewidth]{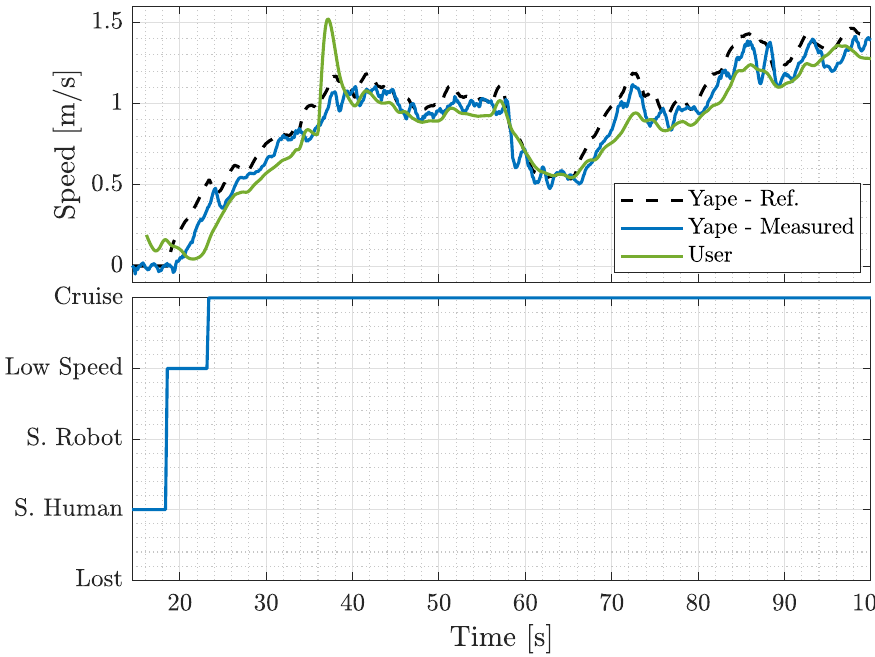}%
  \caption{Supervisor FSM state in relation with the robot’s and user velocity during the starting phase of the experiment of Figure \ref{fig:niguarda_validation/entrance_to_cardio_distance_ctrl}.}
  \label{fig:niguarda_validation/entrance_to_cardio_fsm}
\end{figure}

Figure \ref{fig:niguarda_validation/entrance_to_cardio_distance_ctrl} shows the robot-user distance, the user and robot velocities as well as the force measured by the smart tether's load cell during a navigation experiment from the entrance building to the cardiology ward.
Overall, the robot manages to maintain a suitable distance from the user, close to the setpoint.
Note that this controller does not have stringent tracking requirements.
Rather, its goal is to maintain the error bounded and close to zero, so that the tether is not forced to spool too much wire inwards or outwards.
The force experienced by the user on the handle never exceeds $1$ kgf in this experiment, testifying that the Smart Tether System controller is working correctly.
Figure \ref{fig:niguarda_validation/entrance_to_cardio_fsm} presents a zoom on the initial instants of the trip, highlighting that no oscillations occurred during the starting phase, thanks to the ad-hoc starting phase implemented through the $LOW~SPEED$ discrete state.
After the start-up period, the system works in cruise mode, correctly fusing the velocity references from the distance controller and the DWA algorithms.

\begin{figure*}
  \centering
  \subfloat[]{
    \includegraphics[width=0.48\textwidth,valign=m]{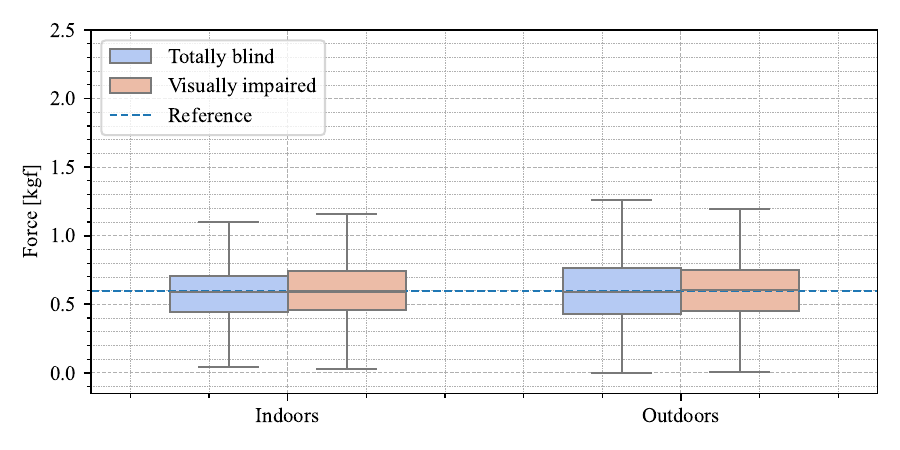}
  \label{fig:niguarda_validation/boxplot_force1}}
  \hfill%
  \subfloat[]{
    \includegraphics[width=0.48\textwidth,valign=m]{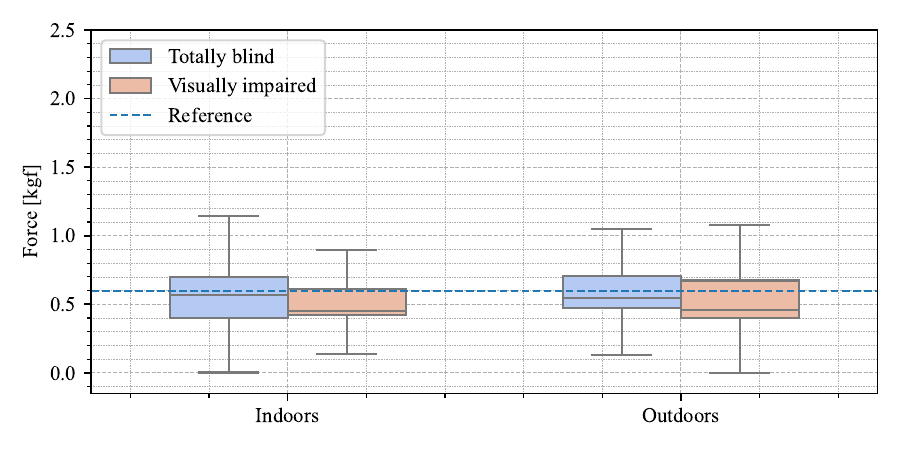}
  \label{fig:niguarda_validation/boxplot_force2}}
  \hfill%
  \subfloat[]{
    \includegraphics[width=0.48\textwidth,valign=m]{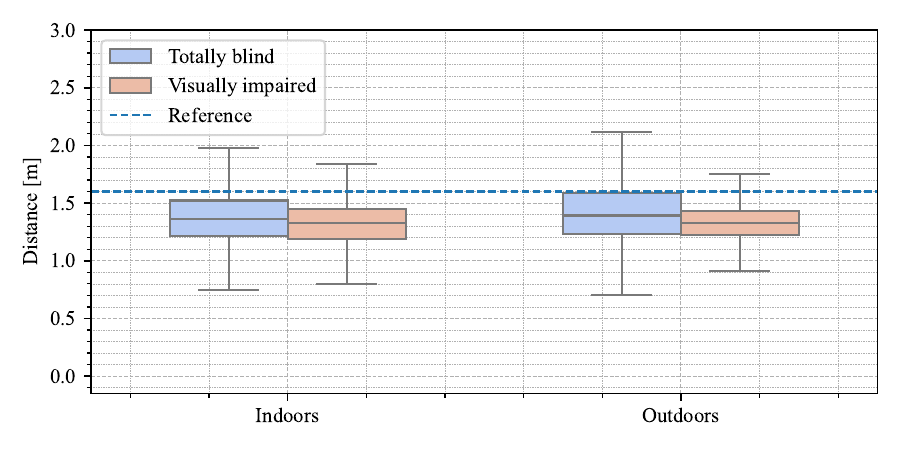}
  \label{fig:niguarda_validation/boxplot_distance}}
  \caption{Distribution of handle force from only the normal state (a), from both normal and stop states (b), and of robot-user distance (c) during the navigation.}
  \label{fig:niguarda_validation/boxplots}
\end{figure*}

The overall distribution of the robot-user distance and the force at the handle during all the tests conducted in the final demo day are reported in Figure \ref{fig:niguarda_validation/boxplots}.
The data are categorized by user type, distinguishing between totally blind and visually impaired individuals, and further separated by navigation environment, both indoors and outdoors. This structure allows a more nuanced evaluation of the system's performance across user profiles and contexts. In total, three plots are provided: one showing the handle force distribution during the normal operating state of the smart tether (Figure \ref{fig:niguarda_validation/boxplot_force1}), another aggregating data from both normal and stop states (Figure \ref{fig:niguarda_validation/boxplot_force2}), and a third illustrating the distribution of robot-user distance across all scenarios (Figure \ref{fig:niguarda_validation/boxplot_distance}). The stop state is typically triggered when the robot halts due to control constraints or user behavior, offering insight into interactions beyond steady-state tracking. 

Across all tested conditions, the controller maintains satisfactory tracking performance, with handle forces and robot-user distances generally aligned with their respective reference targets. In particular, Figure \ref{fig:niguarda_validation/boxplot_force1} demonstrates a strong consistency with the desired handle force, indicating that during normal operation, both user groups interact with the robot in a stable and predictable manner. Table \ref{pvalue} supports this observation in which the force's p-values from both indoor and outdoor are greater than 0.01, showcasing a consistent behavior for both groups. However, behavior differences become evident in Figure \ref{fig:niguarda_validation/boxplot_force2}, where the inclusion of stop events reveals a divergence between user types. Visually impaired users tend to exhibit lower applied forces during these combined states. Their partial visual perception allows them to better anticipate the robot’s trajectory and environmental context, which reduces their reliance on haptic feedback through the tether.

This anticipatory behavior also contributes to a slight reduction in the robot-user distance with respect to the reference distance, as seen in Figure \ref{fig:niguarda_validation/boxplot_distance}. This deviation appears more prominent among visually impaired users. Since the visually impaired users are keeping up with or even slightly overtaking the robot rather than following passively, they often take the lead, walking at a quicker pace and reducing the effective gap between themselves and the robot. This effect is even more visible in outdoor environments, where open space and the absence of close obstacles allow users to move more freely and confidently. Feeling safe and unrestricted, they naturally increase their walking speed, causing the robot to reach its saturation velocity, which is capped at 1.5 meters per second for safety and comfort. Once this upper speed limit is reached, the robot can no longer accelerate to increase the gap, resulting in persistently lower handle forces and shorter distances. Table \ref{pvalue} supports this by demonstrating very small p-values for forces in both normal and stop states and distances.


\begin{table}[H]
	\centering
	\begin{tabular}{|l|l|l|}
		\hline
		& \textbf{Indoor} & \textbf{Outdoor} \\
		\hline
		Force in only normal states & 0.132 & 0.121\\ \hline
		Force in normal and stop states & 0.000 & 0.000\\ \hline
		Distance & 0.000 & 0.000\\ \hline
	\end{tabular}
	\caption{P-values for totally blind and visually impaired (we write p-value smaller than 0.0001 as 0.000)}
	\label{pvalue}
\end{table}

\section{Conclusions}\label{sec:conclusions}
In this work we have described the design and the realization of a prototype of the novel guide robot BUDD-e for blind and visually impaired users. The robot has been tested in a real scenario with the help of visually disabled volunteers at ASST Grande Ospedale Metropolitano Niguarda, in Milan displaying promising performance and results. To assess this, some important feedbacks were received by the volunteers that performed the tests. In general, the users felt that, with BUDD-e, it is possible to navigate easily in places unknown by the user and to walk at significant walking speed. They stated that the users can understand the indications and the route to follow without difficulties, similarly to what happens with a human guide; they finally remarked that BUDD-e does not make unexpected autonomous decisions or distracts, contrarily to what the guide dog sometimes does.\smallskip\\
Notice that, since the robot described in this work is a prototype, current and future research work is dedicated to provide further functionalities arised from user feedbacks and our observations during the tests. In this respect, constructive criticisms were raised from the users, that will lead our future work on this project.\\
First of all, some users suggested to provide more audible vocal indications through the loudskeaker and possibly increase the number of information that it provides, e.g., location, the distance or time that remains to reach the final point, floor levels and materials (similarly to the information that one can receive using a white cane) or alerts about the arrival at destination.\\
Secondly, the need of a secondary interaction with the robot was raised, possibly though vocal commands or a keyboard to give instructions, to modify the route, or to contact the staff in case of issues.\\
From the structural point of view, some volunteers claimed that a reduction of the actual dimensions could make the robot more agile, especially in narrow spaces. Regarding the Smart Tether System, some users required adjust the reference force during curves and  according to the user's preferences.\smallskip\\
Finally, some users pointed out that BUDD-e's functionalities should be made adaptable based on the level of disability. This is perfectly consistent with the remarks done in Section~\ref{sec:user tests conclusions}: BUDD-e is apparently used in a smoother and with less difficulties by blind volunteers with respect to partially sighted users. This, as discussed, is probably due to the fact that the former rely entirely on BUDD-e for assistance, while visually impaired volunteers who retained some partial vision relied less on BUDD-e and on the force exterted through the Smart Tether System, making it less effective.\\
A further remark on the results achieved during the tests is in order. Especially when the environment was particularly crowded, we noted that the surrounding people appeared largely indifferent to BUDD-e's presence, to the point of walking alongside or against it, in one particular case even by touching it. On the one hand, this response suggests a positive social integration of BUDD-e, but future reasearch needs to be specifically devoted to prevent possible issues arising from this fact.
\section*{Acknowledgments}
The authors would like to thank Unione Italiana dei Ciechi e degli Ipovedenti ONLUS-APS, Fondazione Istituto dei Ciechi di Milano ONLUS, Real Eyes Sport, and ASST Grande Ospedale Metropolitano Niguarda for the support and the fruitful collaboration during the project.
%
%
\bibliographystyle{IEEEtran}
\bibliography{biblio,Thesis_bibliography}  
%
%
%
%
\end{document}